\documentclass{article}

\usepackage[preprint]{neurips_2025}
\usepackage[T1]{fontenc}
\usepackage{lmodern}

\usepackage{amsmath,amssymb,amsfonts}
\usepackage{booktabs}
\usepackage{array}
\usepackage{tabularx}
\usepackage{multirow}
\usepackage{longtable}

\usepackage{graphicx}
\usepackage{float}      
\usepackage{placeins}   
\usepackage{subcaption}
\usepackage{wrapfig}

\usepackage{microtype}
\usepackage{nicefrac}
\usepackage{xspace}
\usepackage{tcolorbox}
\usepackage{soul}
\usepackage{enumitem}
\usepackage{listings}
\usepackage[table,dvipsnames]{xcolor}
\usepackage{url} 
\usepackage[colorlinks=true, linkcolor=black, citecolor=black, urlcolor=black]{hyperref}

\lstdefinestyle{mystyle}{
  breaklines=true,
  basicstyle=\ttfamily\small,
  frame=tb,
  xleftmargin=0.05\textwidth,
  xrightmargin=0.05\textwidth,
  columns=flexible,
  keepspaces=true,
  backgroundcolor=\color{gray!10}
}

\newcommand{\hlorange}[1]{\sethlcolor{orange!30}\hl{#1}}
\newcommand{\hlyellow}[1]{\sethlcolor{yellow!30}\hl{#1}}
\newcommand{\hlred}[1]{\sethlcolor{Red!25}\hl{#1}}
\newcommand{\hlblue}[1]{\sethlcolor{RoyalBlue!25}\hl{#1}}
\newcommand{\hlteal}[1]{\sethlcolor{Cyan!25}\hl{#1}}
\newcommand{\hlgreen}[1]{\sethlcolor{Green!25}\hl{#1}}
\newcommand{\hlpurple}[1]{\sethlcolor{Plum!25}\hl{#1}}

\newcommand{\draftcomment}[3]{{\color{#2}[\textit{#1}]$_\text{#3}$}}
\renewcommand{\draftcomment}[3]{}

\newcommand{\qinyuan}[1]{\draftcomment{#1}{teal!60}{QY}}
\newcommand{\risha}[1]{\draftcomment{#1}{purple}{RS}}

\newcommand{\framework}{{\fontfamily{lmss}\selectfont ChEmREF}\xspace}
\newcommand{\hazmat}{\textsc{HazMat}\xspace}

\newcommand{\unnumber}{\hlorange{UN-Number}\xspace}
\newcommand{\commonname}{\hlred{Common Name}\xspace}
\newcommand{\synonyms}{\hlyellow{Synonyms}\xspace}
\newcommand{\molecularformula}{\hlteal{Molecular Formula}\xspace}
\newcommand{\iupacname}{\hlgreen{IUPAC Name}\xspace}
\newcommand{\inchi}{\hlblue{InChI}\xspace}
\newcommand{\smiles}{\hlpurple{SMILES}\xspace}

\title{ChEmREF: Evaluating Language Model Readiness for Chemical Emergency Response}

\author{Risha Surana \quad  Qinyuan Ye \quad Swabha Swayamdipta\\
  University of Southern California\\
  \texttt{\{rsurana,qinyuany,swabhas\}@usc.edu}}

\begin{document}

\maketitle

\begin{abstract}
Emergency responders managing hazardous material (\hazmat) incidents face critical, time-sensitive decisions, manually navigating extensive chemical guidelines.
We investigate whether today's language models can assist responders by rapidly and reliably understanding critical information, identifying hazards, and providing recommendations.
We introduce the Chemical Emergency Response Evaluation Framework (\framework), a new benchmark comprising questions on 1,035 \hazmat chemicals from the Emergency Response Guidebook and the PubChem Database. 
\framework is organized into three tasks: 
(1) translation of chemical representation between structured and unstructured forms (\textit{e.g.}, converting ``C$_2$H$_6$O'' to ``ethanol''), (2) emergency response generation (\textit{e.g.}, recommending appropriate evacuation distances) and (3) domain knowledge question answering from chemical safety and certification exams.
Our best evaluated models received an exact match of 68.0\% on unstructured \hazmat chemical representation translation, a LLM Judge score of 52.7\% on incident response recommendations, and a multiple-choice accuracy of 63.9\% on \hazmat examinations.
These findings suggest that while language models show potential to assist emergency responders in various tasks, they require careful human oversight due to their current limitations.

\end{abstract}

\section{Introduction}

In a hazardous material (\hazmat) incident, ``the decisions made and actions taken in the first few minutes of a response will often establish the character of the overall response – and ultimately its success or failure~\citep{chemm_detailed_info}.''
During the critical ``golden hour,'' first responders must quickly assess the nature and scale of the incident, establish safety protocols, and request resources while protecting themselves and the public.
In current practice, first responders rely extensively on the \hazmat Emergency Response Guidebook (\textbf{ERG}; \citealp{phmsa_erg})\footnote{See \S\ref{appendix:glossary} for a glossary of important terms related to HAZMAT and emergency
response.}: a printed manual of 400 pages that links common \hazmat chemicals with their corresponding incident response guidelines. 
A major challenge in emergency response is navigating the extensive content of the ERG quickly and accurately to identify the appropriate response measures.
In addition, first responders are often required to draw inferences from environmental cues, placards, databases, container labels, victim symptoms, and other contextual information. This adds complexity to the task, demanding skills and interdisciplinary knowledge that were not explicitly documented in the ERG.

Language models (LMs) today have acquired extensive knowledge through large scale pre-training and can synthesize vast amounts of information to generate responses that adhere to complex and dynamic contexts.
Given this progress, we ask: are language models ready to assist chemical emergency responders in \hazmat incidents by automating parts of their workflow?
We believe that the broad knowledge capacity, strong information synthesize ability and accessible interface of LMs open up new possibilities for supporting this task. Additionally, their fast information processing speed may help save valuable time in high-pressure, time-critical decision-making scenarios.

\begin{figure}[t!]
    \centering
    \includegraphics[width=\columnwidth]{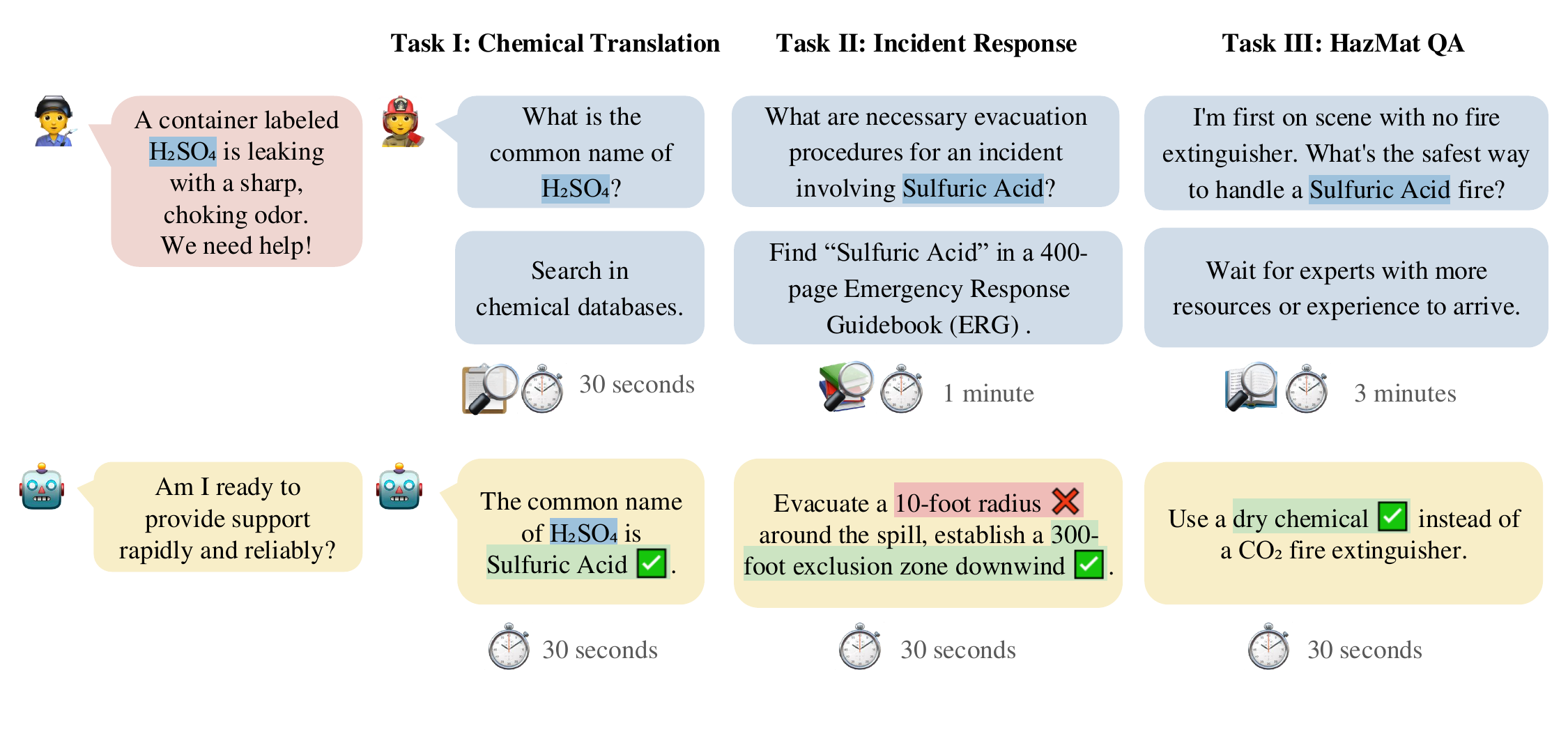}
    \vspace{-0.8cm}
    \caption{\textbf{LLM Assistants in Emergency Response.} In \hazmat emergencies, timely decisions during scene size-up are critical \citep{scenesizeup}. This figure compares the traditional chemical response timeline with one aided by an LLM. While LLMs may accelerate decision-making, they can also produce incorrect guidance. These dynamics informed the design of \framework.
    }
    \label{fig:intro}
    \vspace{-0.2cm}
\end{figure}

To this end, we introduce the \textbf{Chemical Emergency Response Evaluation Framework} (\framework), designed to systematically assess the capabilities of language models in assisting with a \hazmat incident. 
\framework is organized into three tasks which reflect the key steps needed in \hazmat emegency response: (I) chemical identification via translating between structured and unstructured representations, (II) decision-making via emergency response generation, and (III) \hazmat domain knowledge question answering. 
Figure~\ref{fig:intro} provides an overview of the tasks and their roles in the emergency response workflow.
By covering 1,035 \hazmat chemicals, 8 chemical representation types, 6 emergency response dimensions and 6 \hazmat exam categories, \framework provides a comprehensive evaluation of LMs in this high-stakes domain.

We evaluate four general-purpose and two domain-specific language models on \framework. Overall, \textsc{GPT-4o} \citep{hurst2024gpt} is the top-performing model among evaluated models, followed closely by \textsc{Med42 (70B)} \citep{christophe2024med42evaluatingfinetuning} and \textsc{LLaMA-3.1 (70B)} \citep{grattafiori2024llama3herdmodels}. Our results show that each model exhibits distinct strengths and weaknesses on \framework. For example, although \textsc{Phi-4} \citep{abdin2024phi4technicalreport} does not lead overall, it achieves a near-perfect exact match (EM) of 97.2\% on structured chemical translation. We also find that, contrary to our expectations, domain-specific models like \textsc{Med42 (70B)} and \textsc{ChemLLM (7B)} \citep{zhang2024chemllmchemicallargelanguage} do not always outperform general-purpose models. This highlights the interdisciplinary nature of chemical emergency response and suggests that task-specific adaptation may be necessary to further improve performance.

We conduct further analysis of each of the three tasks in \framework to better understand common failure modes and derive actionable insights.
In Task I, we find that models perform significantly better on translations between structured chemical representations (91.7\% EM averaged across models) than on translations involving unstructured representations (60.5\% EM).
This reveals a notable gap: while structured formats are more easily processed by language models, unstructured representations are more commonly used in human communication during \hazmat incidents.
Future systems should account for this gap and better align models with how people communicate in real-world scenarios.
In Task II, our analysis reveals that models struggle to provide emergency response recommendations involving numerical values, often neglecting to include them or significantly overestimating the quantities. Additionally, model outputs are frequently overly verbose and go beyond what is asked, which can be problematic in time-sensitive scenarios.
Finally, all evaluated models struggle in Task III: \hazmat exams, with \textsc{GPT-4o} achieving the highest accuracy of 63.9\%, which is insufficient for high-stakes applications. Breaking down performance by exam category, we find that models perform better on instructional topics such as Hazard Communication \& Workplace Safety, but struggle with more technical and risk-focused topics such as CBRN \& Chemical Safety. These results indicate that current models lack critical domain knowledge that are necessary for \hazmat emergency response.


Overall, our results suggest that language models show potential to assist first responders in \hazmat emergencies, but must be used with great caution and under the supervision of trained professionals, given the high-stakes nature of these situations.
We will publicly release the data and code for \framework and hope that future work will build on it to advance research in this critical domain.


\section{Related Work}
\label{sec:related}
\paragraph{Evaluating LMs on Chemistry Knowledge.} 
Language technologies have long been applied to the chemistry domain, with \citet{thorne-akhondi-2024-nlp} providing a comprehensive review. Recent benchmarks like ChemLLMBench \citep{guo2023largelanguagemodelschemistry} and ChemBench  \citep{mirza2024largelanguagemodelssuperhuman} evaluate large language models (LLMs) on a range of chemistry knowledge and skills, demonstrating their potential while also highlighting limitations such as difficulties with \smiles 
~representations, hallucinations, and overconfident predictions.
Complementary to these works, our work offers a distinct perspective by focusing on the critical and specific domain of \hazmat chemicals. Furthermore, we move beyond memorizing chemicals properties to actively reasoning about and applying them in real-world \hazmat emergency response.
\paragraph{Adapting and Augmenting LMs for Chemistry.} 
While general-purpose LMs show promise in chemistry, further specialization is possible through techniques like continued pre-training or instruction tuning \citep{zhang2024chemllmchemicallargelanguage,christofidellis2023unifyingmoleculartextualrepresentations}.
Separately, researchers have explored incorporating molecular structure as a distinct modality~\citep{edwards-etal-2022-translation} with contrastive learning~\citep{liu2023multi}, or augmenting general-purpose LMs with chemistry-specific tools~\citep{m2024augmenting}. \citet{han-etal-2025-generalist} provide a recent overview of this area.
In this work, we focus on single-modality LMs due to their prevalence. We benchmark both general-purpose LMs and domain-specialized LMs to investigate the impact of domain-specific training on \hazmat-related tasks.
\paragraph{NLP for Emergency Response.} The NLP community has a history of exploratory contributions to various aspects of emergency response.
For example, \citet{watanabe-etal-2013-computer} developed a system to automatically extract key information from emergency management logs for database construction. 
\citet{katsakioris-etal-2021-learning} investigated the problem of converting natural language descriptions of an incident location into GPS coordinates. 
Other work, such as \citet{anikina-2023-towards}, has focused on dialogue act classification and slot tagging for emergency response dialogues.  
These works demonstrate the potential of language technology in the domain of emergency response. 
With the advent of large language models, we revisit this potential and explore their applicability to \hazmat emergency response.

\paragraph{Extended Discussion.} For an extended discussion of related work, refer to Appendix \ref{appendix:extended-related}.

\section{\framework: Chemical Emergency Response Evaluation Framework}
\label{sec:framework}
We introduce \framework: a framework for evaluating whether LMs have the capabilities to assist first responders in chemical emergencies.
Our framework consists of three tasks: (1) Chemical Representation Translation, which tests whether models can accurately convert between different unstructured (\textit{e.g.}, \unnumber,
\commonname, and \synonyms) or structured (\textit{e.g.}, \molecularformula, \inchi, \smiles, and \iupacname\footnote{
\iupacname is set by the International Union of Pure and Applied Chemistry. \inchi~refers to International Chemical Identifier. \smiles~refers to Simplified Molecular Input Line Entry System.}
) chemical representations (see Table~\ref{tab:chem-repr}); (2) Incident Response Recommendation, which measures the ability of LMs to generate relevant safety measures, evacuation distances, and containment strategies based on incident context; and (3) \hazmat Examination, which assesses performance on general-knowledge multiple-choice questions from real-world certification exams for first responders and chemical safety tests.  We present abbreviated examples of each tasks in Table~\ref{tab:translation_ex}. In the following, we describe the background, task formulation, data collection process, and evaluation protocol for each task.

\begin{table}[t]
\scalebox{0.85}
{
\begin{tabular}{p{1.1\columnwidth}}
\toprule
\textbf{Task I: Chemical Representation Translation (\S\ref{ssec:translation_task})}\\\midrule
\textit{Question:} What is the SMILES format for C\textsubscript{4}H\textsubscript{10}O? 

\textit{Answer:} {\color{RoyalBlue} CC(C)(C)O}
\\\midrule
\textbf{Task II: Emergency Incident Response (\S\ref{ssec:incident-response})}\\\midrule
\textit{Scenario:} A leak of \textit{chlorine gas} has occurred near a populated area. Describe immediate public safety and evacuation steps necessary to manage the situation effectively. 

\textit{Recommendation:} 
{\color{RoyalBlue} Stage emergency vehicles 300 feet away from the edge of the spill site... }
\\\midrule
\textbf{Task III: \hazmat Examination (\S\ref{ssec:certification})}\\\midrule
\textit{Question:} When transporting hazardous materials, when must the driver check the vehicle’s tires?

\textit{Answer Choices:}
A) Only at the start of the trip.  
B) Every 150 miles or every three hours, whichever comes first.  
C) Only if the tire pressure warning light turns on.  
D) After reaching the destination.  

\textit{Answer: }{\color{RoyalBlue}B) Every 150 miles or every three hours, whichever comes first.}\\
\bottomrule
\end{tabular}
} 
\vspace{0.2cm}
\caption{\textbf{Overview of Three Core Tasks in \framework{} (Section~\ref{sec:framework}).} 
This table provides representative examples for each of the three core tasks evaluated in our framework. 
}
\vspace{-0.5cm}
\label{tab:translation_ex}
\end{table}



\subsection{Task I: Chemical Representation Translation}
\label{ssec:translation_task}
\begin{table}[ht!]
\centering
\scalebox{0.8}{
\begin{tabular}{l|l|l|l}
\toprule
\multicolumn{2}{c|}{\textbf{Unstructured Representations}} & \multicolumn{2}{c}{\textbf{Structured Representations}}\\
\midrule
\hlred{Common Name}    & tert-Butyl alcohol & \hlgreen{IUPAC Name}       & 2-methylpropan-2-ol\\ 
\hlorange{UN-Number}      & 1120 & \hlteal{Molecular Formula}  & C\textsubscript{4}H\textsubscript{10}O \\
\hlyellow{Synonyms}          & Trimethylcarbinol & \hlblue{InChI}           & InChI=1S/C\textsubscript{4}H\textsubscript{10}O/c1-4(2,3)5/h5H,1-3H3\\
& 2-Methyl-2-propanol & \hlpurple{SMILES}         & CC(C)(C)O  \\ \bottomrule


\end{tabular}
}
\vspace{0.5cm}
\caption{\textbf{Different Chemical Representations of tert-Butyl alcohol, a sample hazardous material.} 
In Task I: Chemical Representation Translation (\S\ref{ssec:translation_task}) we evaluate LMs on translating between these representations, with a focus on translating to/from \hlred{Common Name}, which is most frequently used in chemical emergency guidebooks.}
\vspace{-2pt}
\label{tab:chem-repr}
\end{table}


On receiving a dispatch call, first responders must rapidly and accurately identify the emergency chemicals involved, often based on descriptions of placards, container labels, and other cues. 
However this process can be complicated by the existence of multiple representations for the same chemical.
In Table~\ref{tab:chem-repr}, we provide 7 different representations for the \hazmat chemical named \textit{tert-Butyl alcohol}. In practice, first responders may spend a few minutes consulting the ERG or electronic databases to obtain the representation required for subsequent decision-making. Task I investigates whether LLMs can assist in this critical task at a fast speed.

To build this task, we first subsample 100 random chemicals from the 1,035 hazardous materials listed in the 2024 Emergency Response Guidebook (ERG) and collect their corresponding chemical representations by cross-referencing the PubChem database. For each chemical, we consider both structured and unstructured translation settings, resulting in 19 source-target translation pairs in total.
\begin{itemize}[leftmargin=*]
    \item \textbf{Structured translation} (12 pairs) involves bi-directional, pairwise translation among the 4 structured representations (\iupacname, \molecularformula, \inchi~and \smiles).
     This setting evaluates whether LLMs can reason over the structure of hazardous chemicals.
    \item \textbf{Unstructured translation} (7 pairs) focuses on translating to/from \commonname, as it is most frequently used in emergency communication. This includes translations between $\{$\unnumber,~\iupacname,~\hlteal{Molecular Formula}$\}\leftrightarrows$~\commonname (6 pairs) and \synonyms $\rightarrow$ \commonname (1 pair).
\end{itemize}

We evaluate model performance using exact match (EM). A prediction receives full credit when the output matches the target chemical representation exactly, and zero otherwise.

\subsection{Task II: Incident Response Recommendation}
\label{ssec:incident-response}
Once the chemical involved in the incident has been identified and the suitable chemical representation is found (Task I), emergency responders must analyze the incident and provide appropriate safety recommendations.
For example, in the event of an anhydrous ammonia leak (commonly referred to as ``ammonia gas''), the responder may advice on safe evacuation distances (\textit{e.g.}, 330 feet), protective gear (\textit{e.g.}, self-contained breathing apparatus, or SCBA) and containment procedures (\textit{e.g.}, applying water spray to reduce vapors).
Task II evaluates whether LMs can assist first responders in drafting emergency response recommendations. 


To construct this task, we extract the official recommendations from the 2024 Emergency Response Guidebook (ERG; ~\citealt{phmsa_erg}). 
For each chemical, the ERG provides detailed guidance across six key dimensions: fire or explosion, health, public safety, protective clothing, spill or leak, and first aid. 
Following this, we prompt LLMs to generate recommendations for each of these six aspects through separate queries. 
For efficient evaluation, we randomly subsample 100 chemicals across the 4 structured representation types from the 1,035 chemicals in Task I, resulting in a total of 2,400 queries for Task II.


For evaluation, we compare the model generated recommendations and the ground-truth recommendations in the ERG using two metrics: \textbf{(1) LLM Judge}~\citep{llm_as_a_judge}: We use GPT-4-Turbo as an automatic evaluator for the quality of each generation. It assigns either \textbf{Incorrect (0)}, \textbf{Partially Correct (0.5)}, or \textbf{Perfect (1)} to each generation. Details on the judging prompts and setup can be found in the \S\ref{appendix:llm-judge-setup}. \textbf{(2) BERTScore-F1}~\citep{zhang2020bertscoreevaluatingtextgeneration}. We use BERTScore-F1 as a secondary metric to capture semantic similarity and surface-level variation in phrasing. We also present a human evaluation on the model outputs as outlined in \autoref{appendix:t2-results}.

We introduce a separate metric named \textbf{(3) Cautiousness} for queries involving numerical outputs such as length, time, and volume (\textit{e.g.}, recommending a \textbf{330}-foot evacuation distance, recommending not to enter the accident site for \textbf{24} hours). This metric quantifies how closely model-generated values align with the ERG. Specifically, we compute an Mean Absolute Relative Error (MARE) metric between the predicted ($\hat{y}_i$) and ground truth ($y_i$) values. 

\[
\text{MARE} \;=\; \frac{1}{n} \sum_{i=1}^{n} \left| \frac{y_i - \hat{y}_i}{y_i} \right|
      \;=\; \frac{1}{n} \sum_{i=1}^{n} \left| \frac{e_i}{y_i} \right|.
\]

\subsection{Task III: \hazmat Examination}
\label{ssec:certification}

Beyond translation and emergency response recommendation, a fundamental understanding of \hazmat concepts is also essential for first responders. Such knowledge can be evaluated through question-answering tasks that more closely reflect real-world scenarios.
In practice, first responders, along with other professionals like lab technicians, truck drivers, and warehouse managers, are required to pass \hazmat certification exams. These exams offer a valuable benchmark for evaluating a model's understanding of \hazmat chemicals.

In Task III, we collect a total of \textbf{865 multiple-choice questions} from 46 publicly available quizzes on ProProfs.\footnote{\url{https://www.proprofs.com/quiz-school/}} These quizzes cover a wide range of topics, including \hazmat Awareness,  Workplace Safety, Transportation Safety and more. 
Each quiz provides answer keys and explanations that have been reviewed by both educators and learners preparing for the exam. Additionally, we manually filtered and verified the questions to ensure quality.
We evaluate LLMs on these questions and report standard accuracy.



\section{Evaluating LMs on \framework}
\label{sec:results}
In this section, we first introduce experiment details such as compared models and prompting strategies in \S\ref{ssec:exp_details}. We then report model performance on \framework and summarize our key observations in \S\ref{ssec:cross-task-observation}. 
In \S\ref{sec:analysis}, we provide more in-depth analyses of each task.

\subsection{Experiment Details}
\label{ssec:exp_details}
\paragraph{Models.} We evaluate six different open-weight LMs on \framework. 
Three are general-purpose models covering a range of scales: \textsc{Phi-3} (3.8B) \citep{abdin2024phi3technicalreporthighly}, \textsc{Phi-4} (14B) \citep{abdin2024phi4technicalreport}, and \textsc{Llama-3.1} (70B) \citep{grattafiori2024llama3herdmodels}. 
We also include two domain-specialized models: \textsc{ChemLLM} (7B) \citep{zhang2024chemllmchemicallargelanguage} for chemistry and \textsc{Med42} (70B) \citep{christophe2024med42evaluatingfinetuning} for medicine. 
Both models are adapted via supervised fine-tuning (SFT), with \textsc{ChemLLM} fine-tuned from \textsc{Intern-LM-2} (7B) \citep{cai2024internlm2} and \textsc{Med42} fine-tuned from \textsc{LLaMA-2} (70B) \citep{touvron2023llama}.
Lastly, we evaluate \textsc{GPT-4o} \citep{hurst2024gpt} as a representative closed-source model.


\paragraph{Task I: Chemical Representation Translation.}
As described in \S\ref{ssec:translation_task}, Task I comprises two settings: structured and unstructured translation.
In a pilot study (Appendix~\ref{appendix:prompt_type}), we experimented with various prompting strategies, including zero-shot and one-shot prompting, both with and without chain-of-thought reasoning.

For \textbf{structured} translation, we adopt \textbf{one-shot chain-of-thought prompting}, as intermediate reasoning steps often aid in breaking down complex chemical terms (\textit{e.g.}, translating ``sodium'' to ``Na'' and ``superoxide'' to ``O\textsubscript{2}'').
For \textbf{unstructured} translation, we employ \textbf{direct zero-shot prompting}, as we observed no consistent improvement from including demonstrations or reasoning steps. We provide prompt examples in Appendix~\ref{appendix:prompt_sets}.



\paragraph{Task II: Incident Response Recommendation.}
We evaluate models on Task II with  \textbf{direct 0-shot prompting} that queries models to provide recommendations in one of the six key dimensions (\textit{e.g.}, fire or explosion hazards, public safety). We provide an example prompt in Appendix~\ref{appendix:prompt_sets}.

\paragraph{Task III: \hazmat Examination.}
Task III involves answering multiple-choice questions with inputs presented in a standardized format: \textit{Question: ... Answer Choices: ... Answer:}. We use \textbf{direct zero-shot prompting} without examples or intermediate reasoning. 

\subsection{Overall Results}
\label{ssec:cross-task-observation}
\begin{table*}[ht!]
\centering
\resizebox{\textwidth}{!}{
\begin{tabular}{@{}lrrrrrrrr@{}}
    \toprule
    \multirow{2}{*}{\textbf{Model}} 
    & \multicolumn{2}{c}{\textbf{I. Translation}} 
    & \multicolumn{3}{c}{\textbf{II. Incident Response}} 
    & \textbf{III. Exam} & \multirow{2}{*}{\textbf{Avg.} ($\uparrow$)} \\
    
    \cmidrule(lr){2-3} \cmidrule(lr){4-6} \cmidrule(lr){7-7}
    
    & \textbf{Struc. EM} ($\uparrow$)
    & \textbf{Unst. EM} ($\uparrow$) 
    & \textbf{LLM Judge} ($\uparrow$)& \textbf{BERTScore} ($\uparrow$)
    & \textbf{MARE} ($\downarrow$) 
    & \textbf{Acc.} ($\uparrow$) \\
    
    \midrule
    Phi-3 (3.8B)      & \underline{93.7} & 60.0   & 42.7 & 52.0 & 3.8   & 49.0 & 59.5 \\
    ChemLLM (7B)      & 79.9 & 56.8   & 46.3 & 79.7 & 120.0 & 47.3 & 62.0 \\
    Phi-4 (14B)       & \textbf{97.2} & 48.7   & 25.2 & 79.9 & \underline{2.8}   & \underline{60.0} & 62.2 \\
    Med42 (70B)       & 88.7 & 61.9   & \underline{50.8} & \textbf{80.5} & 3.7   & 58.0 & \underline{68.0}\\
    Llama-3.1 (70B)   & 93.1 & \underline{67.3}   & 50.7 & 64.9 & 5.1 & \underline{60.0} & 67.2 \\
    GPT-4o            & 92.4 & \textbf{68.0}   & \textbf{52.7} & \underline{80.2} & \textbf{2.0}   & \textbf{63.9} & \textbf{71.4} \\
    \bottomrule
\end{tabular}
}
\caption{\textbf{\framework Evaluation Results.} \textbf{Bold} and \underline{underlined} values indicate the \textbf{best} and \underline{second-best} performance in each column. EM stands for exact match. Average is computed over all columns except for MARE.
}
\vspace{-0.3cm}
\label{tab:overall_model_summary}
\end{table*}

\paragraph{\textsc{GPT-4o} is the best overall performer, with \textsc{Med42} and \textsc{Llama-3.1} following closely.}
We present the \framework evaluation results in Table~\ref{tab:overall_model_summary}. In addition to performance on individual tasks, we compute an overall average score over all columns (excluding cautiousness MARE) for high-level comparison of different models.
We found that \textsc{GPT-4o} achieves the highest overall average score of 71.4\%, followed by \textsc{Med42} (68.0\%) and \textsc{Llama-3.1} (67.2\%).

\begin{wrapfigure}{r}{0.49\textwidth}
  \centering
  \includegraphics[width=0.37\textwidth]{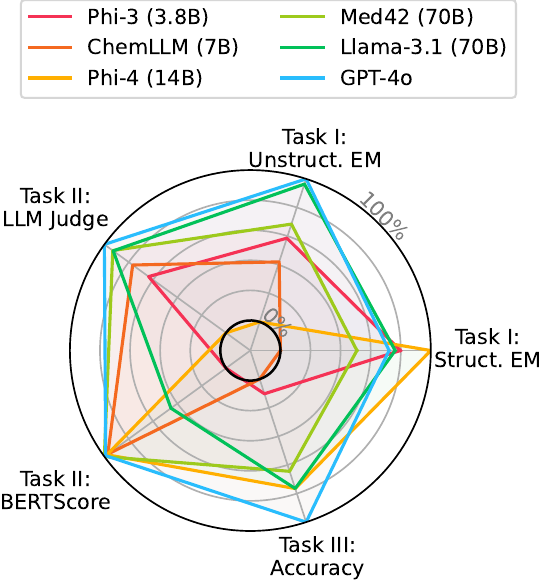}
  \caption{\textbf{\framework Evaluation Results With Per-Metric Normalization.} To facilitate clearer model comparison, we normalize each column in Table~\ref{tab:overall_model_summary} to the [0,1] range. This visualization reveals that models exhibit distinct strengths and weaknesses across tasks.}
  \vspace{-0.5cm}
  \label{fig:radar-results}
\end{wrapfigure}

\paragraph{Models exhibit distinct strengths and weaknesses.}
In Figure~\ref{fig:radar-results}, we visualize the results in Table~\ref{tab:overall_model_summary}, with each metric column normalized to the range of [0,1] to enable clearer model-wise comparison.
We observe that individual models often excel in specific tasks while under-performing in others. For example, \textsc{Phi-4} achieves the highest score in Structured Translation (97.2\% EM) but performs the worst in Incidence Response (25.2\% LLM Judge Score).
These trade-offs manifest as skewed or imbalanced shapes in Figure~\ref{fig:radar-results}. 
While \textsc{GPT-4o}, \textsc{Med42}, and \textsc{Llama-3.1} show promising overall performance, it is important to remain aware of each model’s distinct weaknesses. 
Their outputs should always be used with human oversight to ensure safe and reliable decision-making in emergency response.

\paragraph{Limitations of domain-specific training.}
\textsc{ChemLLM} and \textsc{Med42} are two models we evaluated that have undergone domain-specific supervised fine-tuning (SFT). Our results suggest that such domain-specific training \textit{does not} always guarantee successful application in chemical emergency response. 
Notably, despite being trained on 7 million chemistry-related QA pairs, ChemLLM ranks the worst in structured chemical translation and second-worst in unstructured chemical translation.

Notably, general-purpose models including \textsc{GPT-4o}, \textsc{Llama-3.1} and \textsc{Phi-4} outperform domain-specialized models on Task III: \hazmat Exam.
We hypothesize that this is due to the interdisciplinary nature of the task, as it may require knowledge spanning chemistry, medicine, emergency response, numerical reasoning, and more.
General-purpose models likely benefit from its broader knowledge coverage, whereas domain-specialized models may lose access to such general knowledge after domain-specific training.
These results also highlight the complexity of chemical emergency response and the limitations of current domain-specific training technique. We leave the exploration of more effective domain or task specific training strategies as future work.

\section{Analyzing LLM performance on \framework}
\label{sec:analysis}

In this section, we further break down model performance on the three core tasks and analyze common failure cases and limitations. We highlight several key observations, with additional discussion deferred to Appendix~\ref{appendix:appendix-results}.

\subsection{Analyzing Task I: Chemical Representation Translation}
\label{ssec:t1-results}


\begin{figure}[ht]
    \centering
    \includegraphics[width=0.78\linewidth]{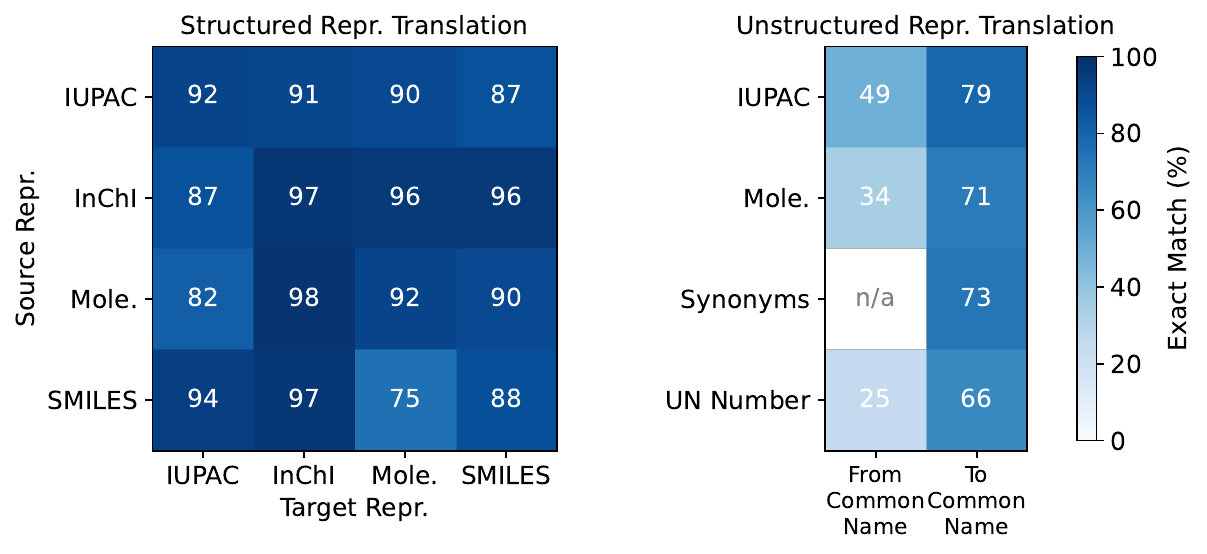}
    \caption{\textbf{Task I Performance Breakdown. Left/Right: Structured/Unstructured Translation.} For brevity, we use ``Mole.'' to denote \molecularformula.
We report EM scores for each source-target pair, averaged across all evaluated models.%
    }
    \label{fig:t1_both_images}
\end{figure}

In Figure~\ref{fig:t1_both_images}, we break down the performance on Task I by source and target chemical representations. The reported performance is averaged with all six evaluated models.

\paragraph{Structured Translation vs. Unstructured Translation.}
We observe that models perform significantly better when translating between \textit{structured} chemical formats such as \iupacname, \molecularformula, \inchi, and \hlpurple{SMILES}, compared to translation involving \textit{unstructured} forms like \hlred{Common Name}, \hlorange{UN-Name}, and \hlyellow{Synonyms}. 
For example, \textsc{Phi-4} achieves a EM of 97.2\% on structured translation, but the performance drops to 48.7\% on unstructured translation. 
Similar trends are observed on other models, highlighting the model's limitation on grounding and normalization when working with less structured or colloquial names.

We have several hypotheses for this performance gap. (1) The structured representations are rule-based and systematic, making them more suitable for step-by-step reasoning through chain-of-thought prompting. In contrast, unstructured translation tends to rely more on the model's memorization than on logical reasoning.
(2) Structured representations are defined by international standards and are likely to appear more frequently in chemical databases and online resources, making them more prevalent in the model's pretraining data. In contrast, unstructured forms like \synonyms are often long-tail, ambiguous, and geographically dependent. 
In general, while unstructured representations are more accessible to humans, structured representations are more machine-actionable. This observation should inform the future design of AI-assisted emergency response systems and effective human-AI collaboration.
\paragraph{To Common Name vs. From Common Name.} 
Translation to \commonname is a critical step in the emergency response workflow, as it often serves as the entry point for subsequent actions such as consulting the Emergency Response Guidebook (see Figure~\ref{fig:intro}).
In Figure~\ref{fig:t1_both_images} (Right), we observe that models perform better when translating to \commonname than from it, which we view as a promising sign.

\subsection{Analyzing Task II: Incident Response Recommendation}
\label{ssec:t2-results}

\paragraph{Evaluating quantitative recommendations and cautiousness.} 

Many emergency response actions rely on quantitative recommendations, \textit{e.g.}, advising an \textbf{330}-foot evacuation radius.
While examining the model outputs in Task II, we first observe that models often neglect to generate quantitative recommendations when expected. 
Among queries where the ground-truth recommendation includes a numerical value, models produce values in the correct category (\textit{e.g.}, length, mass, time, concentration) in only 20\% of cases.

For outputs that include valid numerical values, we further compute the mean absolute relative error (MARE) metric (defined in \S\ref{ssec:incident-response}) as a measurement of model's cautiousness.
\textsc{ChemLLM} demonstrates the most cautious behavior, with an MARE of \textbf{120.03}.
Our manual inspection reveals its tendency to overestimate values, meaning that the quantities recommended by ChemLLM are on average 120 times higher than the groundtruth, which could place unnecessary strain on response efforts.
\textsc{GPT-4o} has the lowest MARE of 2.0, which is significantly lower than 120.0. While overestimation still occurs, it is far less severe and likely more manageable in practical settings.


Our analysis on cautiousness also emphasize the limitations of automatic metrics like LLM Judge or BERTScore. In high-stakes domains like chemical emergency response, it is essential to assess not only the overall plausibility of a recommendation, but also whether it produces a quantitative value when needed, and whether that value is accurate.


\paragraph{Verbose outputs may hinder time-sensitive decision-making.}
In our manual review of 20 recommendations generated by \textsc{GPT-4o}, we find that the model frequently produces outputs that extend beyond the scope of the prompt.
For example (\S\ref{appendix:t2_model_outputs}), when asked specifically about the fire or explosion hazards of a given chemical, \textsc{GPT-4o} additionally generates guidance on personal protective equipment (PPE) and continuous assessment of hazards, resulting in outputs that are longer and less focused.
While such comprehensive responses demonstrate the model's broad knowledge, this verbosity may pose challenges in time-sensitive scenarios, where concise and actionable information is preferred.


\risha{TODO: add human evaluation results + results from MTURK + inner-annotator agreement}

\subsection{Analyzing Task III: \hazmat Examination}
\label{ssec:t3-results}
\begin{table}[ht]
\centering
\scalebox{0.9}{
\begin{tabular}{l c c}
\toprule
\textbf{Category} & \textbf{Average Accuracy} & \textbf{Number of Exams} \\
\midrule
CBRN \& Chemical Safety                     & 46.5 & 3 \\
\hazmat Awareness \& Operations              & 51.9 & 15 \\
\hazmat Lesson Plans                         & \underline{70.8} & 8 \\
Hazard Communication \& Workplace Safety    & \textbf{75.6} & 9 \\
IMDG \& Transportation Safety               & 66.7 & 1 \\
Incident Management \& Reporting            & 62.6 & 10 \\
\midrule
\textbf{Total Exam Cohort}           & 62.0 & 46 \\
\bottomrule
\end{tabular}
}
\vspace{0.5cm}
\caption{\textbf{Task III Accuracy by Category.} We report mean accuracy across all evaluated models. Models perform better on categories with instructional content, but struggle with technical and risk-related topics, highlighting their limitations in these specialized domains.}
\vspace{-0.5cm}

\label{tab:mc-by-exam}
\end{table}


\paragraph{Performance varies across \hazmat exam categories.}  
When gathering \hazmat-related exams from ProProfs, we manually group them into six major exam categories. In Table~\ref{tab:mc-by-exam}, we report the accuracy on different exam categories.
Models perform well on instructional content, such as Hazard Communication and Workplace Safety (Accuracy: 75.6\%) and \hazmat Lesson Plans (Accuracy: 70.8\%), but struggle with more technical and risk-related topics, including CBRN \& Chemical Safety (Accuracy: 46.5\%) and \hazmat Awareness \& Operations (Accuracy: 51.9\%). This discrepancy highlights limitations in  current models’ ability to handle complex, high-stakes topics that require deeper domain understanding.

\section{Conclusion}
\label{sec:conclusion}

In this work, we introduce \framework, an evaluation framework for LLMs covering three tasks related to chemical emergencies. Our results show that while models perform well with structured representations, used in the Task I: Chemical Representation Translation and the Task II: Incident Response Recommendation, they struggle with the chemical representations that are less structured but more commonly used in human colloquial communication. Addressing this human-AI communication gap should be a key consideration in the design of future AI-assisted emergency response systems.
Our results on Task III: Hazmat Examination, suggest that current domain-specific post-training approaches may lead to a loss of general capabilities, making them less suitable for highly interdisciplinary domains like chemical emergency response.
More broadly, we believe LLMs have the potential to assist with chemical identification and incident response, but they should be used with caution and positioned as a complement to, rather than a replacement for, human experts in emergency response.

Future work could extend our \framework benchmark in several important directions.
First, beyond prompting models and evaluating their parametric knowledge, future efforts may explore augmenting models with external retrieval, long-context capabilities, or agentic workflows. Integrating resources such as the full Emergency Response Guidebook (ERG) or chemical databases could enable models to generate more informed, context-aware, and targeted recommendations.
Second, future extensions could incorporate a vision module capable of interpreting cues such as \hazmat placards and container labels, and an audio module for analyzing and engaging with real-time dispatch communications.
Third, conducting real-world studies with domain experts and emergency responders would provide valuable validation of model outputs, help identify critical failure modes, and support future research on human–AI collaboration in high-stakes environments.

\bibliographystyle{plainnat}
\bibliography{custom}

\appendix





\clearpage
\section{Glossary}
\label{appendix:glossary}
\begin{table}[ht]
\scalebox{0.88}{
\begin{tabular}{p{5cm} p{10cm}}

    \toprule
    \textbf{Term} & \textbf{Definition} \\
    \midrule
    Hazardous Materials (\hazmat) & Substances that pose a risk to health, property, or the environment, often requiring special handling and regulations. \\
    \midrule
    Emergency Response Guidebook (ERG) & A resource used by first responders to identify hazardous materials and guide their response during transport incidents. \\
    \midrule
    International Union of Pure and Applied Chemistry (\iupacname) & A global organization that sets standards for chemical nomenclature, terminology, and measurement. \\
    \midrule
    Simplified Molecular Input Line Entry System (\smiles) & A notation system that allows the representation of a molecular structure using a linear string of text. Regarding \smiles evaluation, we rely on the single \smiles string that PubChem now provides for every compound. PubChem has deprecated the separate canonical and isomeric fields in favor of a unified \smiles that is both canonicalized and stereo-/isotope-explicit. For each compound, this hybrid and unique canonicalized form is the isomeric \smiles and was used verbatim during both training and evaluation. As such, the task is designed to reproduce PubChem’s published representation, not to normalize across equivalent non-canonical variants which do not have unique representations (i.e., one non-cannocalized form can represent multiple chemicals). \\
    \midrule
    International Chemical Identifier (InChI) & A textual identifier that provides a unique representation of chemical substances to facilitate data sharing and searchability. \\
    \midrule
    \unnumber & A \unnumber is a four-digit code used to identify hazardous materials for safe transportation and emergency response, regulated by frameworks such as the International Maritime Dangerous Goods (IMDG) Code, International Air Transport Association (IATA) Dangerous Goods Regulations, U.S. Department of Transportation (DOT) Regulations, and the European Agreement concerning the International Carriage of Dangerous Goods by Road (ADR). \\
    \midrule
    Commercial Driver's License (CDL) & A specialized license required for individuals operating large, heavy, or hazardous material-carrying commercial vehicles. \\
    \midrule
    Compound Identifier (CID) & A unique numerical identifier assigned to chemical substances in the PubChem database for reference and research purposes. \\
    \midrule
    Personal Protective Equipment (PPE) & Gear and clothing designed to protect individuals from hazardous materials, contamination, or injury in various environments. \\
    \midrule
    National Registry of Emergency Medical Technicians (NREMT) & A certification organization that ensures emergency medical personnel meet national training and competency standards. \\
    \bottomrule
\end{tabular}
}
\caption{\textbf{Key Definitions.} A glossary of important terms related to \hazmat and emergency response.}
\end{table}

\clearpage

\section{Extended Related Work}
\label{appendix:extended-related}

\paragraph{Existing Software.}
While automated tools have been developed to offer support in emergency response, they are often limited to specific cases or lack integration with the latest language technologies.
For example, CHEMM \citep{chemm_chemical_incidents} provides an automated chemical identification tool based on patient vitals, such as pupil size and skin condition. However, this functionality is limited to later steps in the response sequence and excludes earlier-stage scenarios like dispatch communications \citep{scenesizeup}.
Another tool, CAMEO \citep{epa_cameo}, offers web interfaces with database search, threat zones modeling and incident site mapping, but lacks modern language technologies like flexible keyword matching or a question-answering interface.
To this day, first responders still largely resort to the physical Emergency Response Guidebook (ERG).
LLMs, with their extensive knowledge capacity and user-friendly interfaces, have the potential to bridge this gap and enhance decision-making during these critical moments.

\paragraph{Efforts in CV and Robotics Research Communities.} 
Beyond the scope of natural language processing and language models, fields such as computer vision and robotics also hold promise for improving emergency response.
\citet{naim2021aiagentsemergencyresponse} discuss the integration of AI agents in emergency response applications, emphasizing their potential to enhance situational awareness and decision-making during critical incidents. 
\citet{sharifi2020deephazmathazardousmaterialssign} introduced Deep\hazmat, a deep learning-based robot designed for hazardous materials sign detection and segmentation under constrained computational resources. 
This approach uses convolutional neural networks (CNNs) to identify hazardous materials in complex environments, facilitating faster and more accurate scene assessments. 
Furthermore, \citet{Cai_2020} demonstrated the effectiveness of incorporating geometric prior knowledge into CNN-based planar object detection, which can significantly improve the accuracy of identifying hazardous materials and related objects in emergency scenarios.

\section{Example Prompts and Outputs}
\label{appendix:prompt_sets}

Previously in Table~\ref{tab:translation_ex}, we used redacted examples due to space limit.
In this section, we provide examples of full inputs and expected outputs (marked in {\color{RoyalBlue} blue}) for each of the three tasks in \framework.

\subsection{Translation Task}
\subsubsection{Unstructured Translation}
You are an expert chemist. Question: What is the molecular formula of Sodium superoxide? \\
Answer: {\color{RoyalBlue} NaO$_2$.}

\subsubsection{Structured Translation}
You are an expert chemist. Question: What is the molecular formula of Sodium superoxide? \\
Answer: Let's think step by step. \\
{\color{RoyalBlue}
Sodium (Na) is an alkali metal with a +1 oxidation state. \\
Superoxide is an ion with the formula O$_2^-$ and a -1 charge. \\
To balance the charges, one sodium ion (Na$^+$) will pair with one superoxide ion (O$_2^-$). \\
Therefore, the molecular formula is NaO$_2$.
}


    
    
    

\subsection{Incident Response Task}
Scenario: A fire has started involving Mercuric nitrate (IUPAC-name). \\
Prompt: Provide recommendations on fire control, explosion risks, and necessary evacuation procedures for the substance involved, referencing the hazmat guide where appropriate.\\
Answer: \\
{\color{RoyalBlue}
Fire Control \\
Use water spray or fog to cool the area and prevent fire spread. \\
Avoid direct water streams that may spread contamination. \\
Use dry chemical powder or CO$_2$ for fire suppression.
\\\\
Explosion Risks \\
Mercuric nitrate is a strong oxidizer that can intensify fires. \\
Keep it away from combustible materials.
\\\\
Evacuation Procedures \\
Evacuate all non-essential personnel at least 100 meters away. \\
Ensure emergency responders wear PPE, including self-contained breathing apparatus.
}

\subsection{Hazmat Examination Task}
Reason through this question carefully and provide the answer as the correct multiple-choice letter. Question: What amount of chlorine being transported requires the driver to stop before railroad crossings?\\
A. An amount requiring a Table 2 placard. \\
B. Any amount. \\
C. Any amount when combined with explosives A. \\
D. 1000 lbs or more.\\
Answer: {\color{RoyalBlue} The correct answer is B. Any amount.}

\section{LLM Judge Setup and Evaluation Protocol}
\label{appendix:llm-judge-setup}

To evaluate model-generated responses against official guidelines, we developed an automated evaluation pipeline that uses GPT-4o as an LLM judge. The task involved comparing model responses to official guidance from the Emergency Response Guidebook (ERG) in six categories:

\begin{itemize} \item \textbf{Fire or Explosion Hazard:} Focuses on whether the model correctly identifies flammability, reactivity, or explosiveness. \item \textbf{Health Hazard:} Assesses accuracy in toxicity, corrosiveness, or inhalation danger details. \item \textbf{Public Safety and Evacuation:} Evaluates evacuation distances, containment zones, and timing, considering whether the model is overly cautious or not cautious enough. \item \textbf{Protective Clothing:} Ensures that the model recommends appropriate personal protective equipment (PPE) as indicated in the guide. \item \textbf{Spill or Leak:} Focuses on containment strategies such as diking, neutralization, or absorption. \item \textbf{First Aid:} Evaluates whether the advice aligns with ERG guidance for eye, skin, or inhalation exposure. \end{itemize}

\subsection{Evaluation Prompt Design}

Each evaluation prompt was designed to guide GPT-4o in judging the model's output relative to the ERG recommendations. The system prompt positioned the LLM as a hazardous materials safety expert tasked with verifying whether the model output correctly covered the major information required under the specified category.

\begin{quote}
\texttt{You are a hazardous materials safety expert. Your task is to evaluate a model-generated hazmat safety recommendation against the official Emergency Response Guide (ERG) guidance. Your goal is to check whether the model covers the major information from the ERG under the given category.
It is acceptable if the model includes extra relevant information not in the ERG, as long as it is accurate. Only penalize if something significant from the ERG is missing or if the model includes incorrect information.}

\texttt{Use one of these labels:  }

\begin{itemize}
    \item \texttt{\textbf{Incorrect}: Key points are missing or wrong.}
    \item \texttt{\textbf{Partial}: Some important points are correct, but others are missing or incorrect.}
    \item \texttt{\textbf{Correct}: Most or all important ERG details are present and accurate.}
\end{itemize}
\end{quote}

For each one of the six categories mentioned above, we  supplement the LLM Judge with specific judging instructions. For example, the judging instructions for ``Fire or Explosion Hazard'' category encourages the LLM Judge to prioritize flammability, reactivity, and explosiveness in its assessment.

\clearpage
\section{Extended Results}
\label{appendix:appendix-results}

In this section, we provide supplementary figures and analysis for the Task I: Translation and Task II: Incident Response.

\subsection{Analyzing Task I: Chemical Representation Translation}

\section{Self-translation}
Self-translations (\textit{e.g.}, from \smiles to \smiles) is a simple extension from the translation task.
We additionally evaluate models on self-translation on the four structured representations, as a sanity check. We report the detailed model-specific results in Fig.~\ref{fig:trans_var_main_result}.
\textsc{LLaMA-3.1} achieved the highest EM score (98.6\%), followed closely by \textsc{Phi-4} (97.8\%) and \textsc{Med42} (95.0\%). 
Surprisingly, \textsc{ChemLLM} achieves a lower score of 86.3\% on this task, suggesting a lack of basic understanding of chemical representation types.
We also observe that model rankings on self-translation are consistent with their ranking on structured translation in Table~\ref{tab:overall_model_summary}, suggesting that this simple check can serve as an early-stage evaluation for filtering out less suitable models.

\subsubsection{Prompting Methods}
\label{appendix:prompt_type}

\begin{figure}[ht]
\begin{minipage}{0.48\textwidth}
    \centering
    \includegraphics[width=\textwidth]{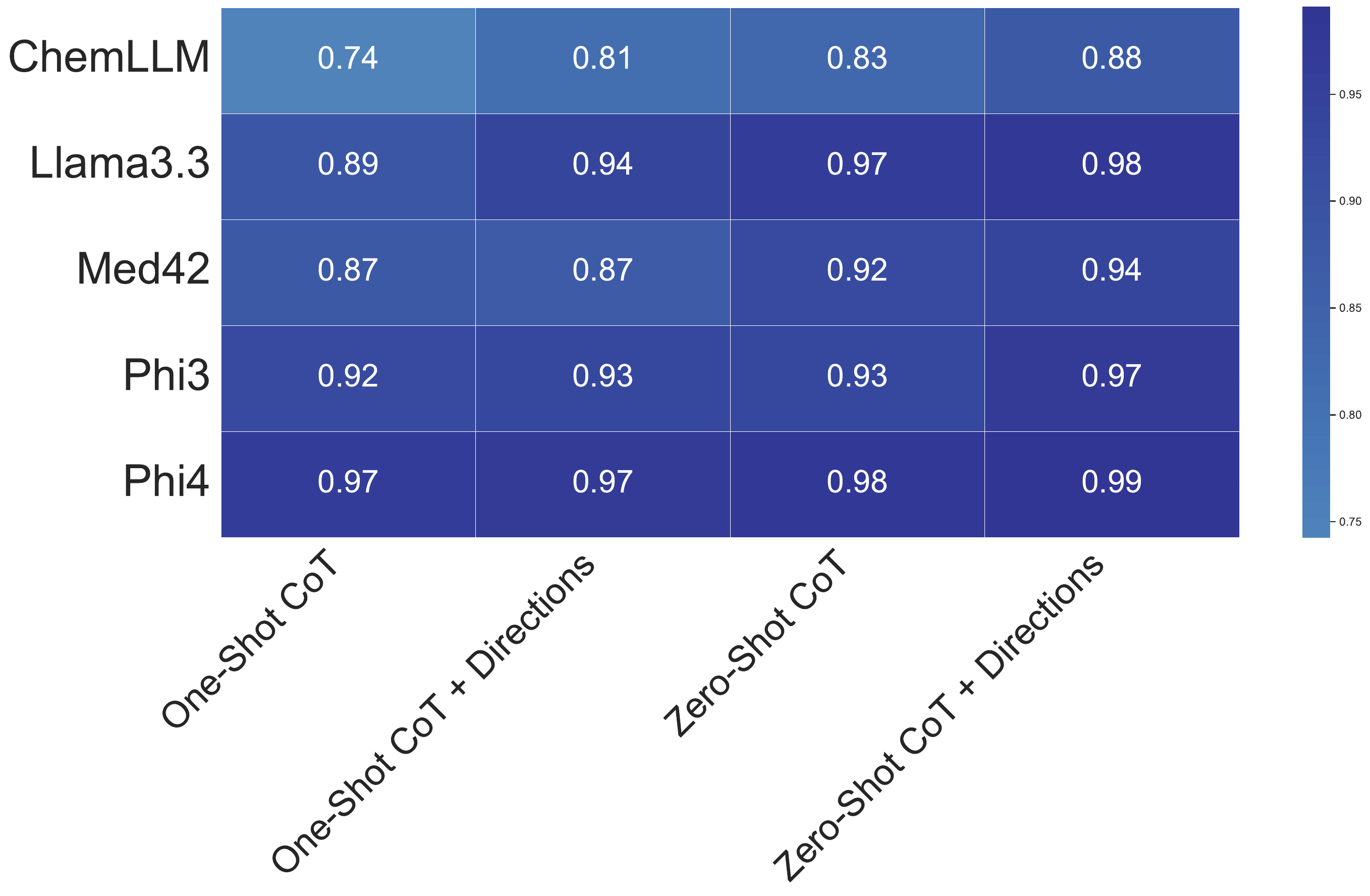}
    \caption{\textbf{Translation Task Model Performance Across Prompt Types:} Performance heatmap using HAZMAT data.}
    \label{fig:translation_by_prompt_hazmat}
    \end{minipage}
\hfill
\begin{minipage}{0.48\textwidth}
\includegraphics[width=\textwidth]{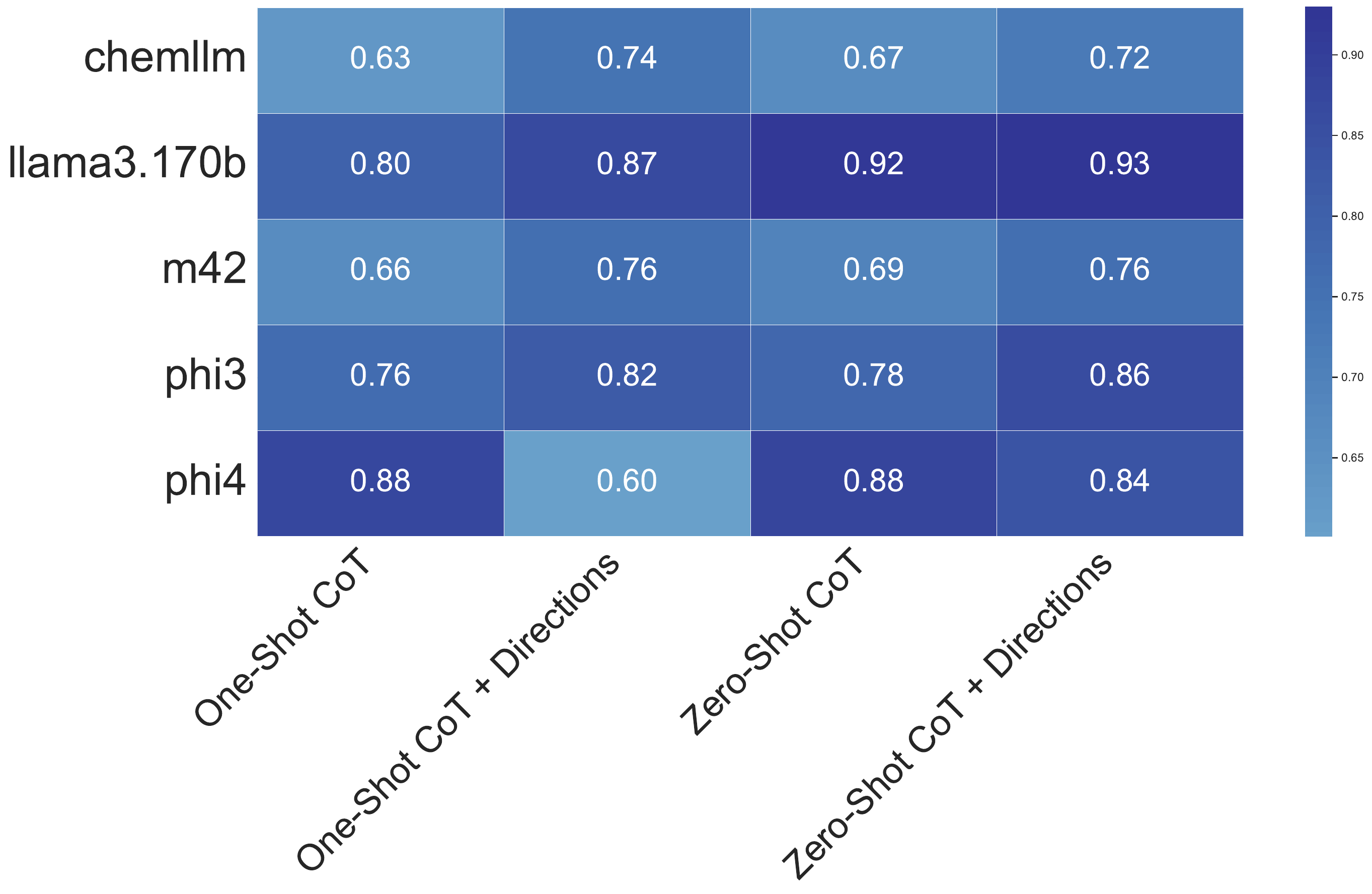}
    \caption{\textbf{Translation Task Model Performance Across Prompt Types:} Performance heatmap using Non-Hazmat chemical data.}
    \label{fig:translation_by_prompt_pub}
        \end{minipage}
\end{figure}

For Task I, we ran a pilot experiment using four different prompting strategies, including (1) zero-shot chain-of-thought (``Let's think step by step''; \citealt{kojima2022large}), (2) zero-shot chain-of-thought with directions, (3) one-shot chain-of-thought prompting \citep{wei2022chain}, and (4) one-shot chain-of-thought prompting with directions.\qinyuan{Did we run the pilot experiments with CoT prompting only? I remember that we have direct prompting experiments too. Why do we choose zero-shot direct prompting for unstructured translation?} \risha{I mention this below!}
Combining the four prompting strategies with the 16 source-target translation pairs by chemical representation type, we experiment with 64 settings for each model for each chemical.
In \autoref{fig:translation_by_prompt_hazmat} and \autoref{fig:translation_by_prompt_pub}, we report the average of the exact match over the 16 source-target pairs.
\paragraph{Performance is slightly sensitive to the prompt format across models; the best prompt format varies with the model.}
As shown in \autoref{fig:translation_by_prompt_hazmat}, the impact of prompt format on performance varies significantly by model.
As we provide the model with less examples, for a model like ChemLLM, prompt format can mean an improved score of up to 14\% on HAZMAT chemicals.
Alternatively, a model like Phi-4 shows minimal performance variation across prompt formats for HAZMAT chemicals, but exhibits significant variation on non-HAZMAT data, achieving performance comparable to ChemLLM on HAZMAT data.
We hypothesize that including too much chemical instruction in the prompt can sometimes lead the model to produce outputs that closely mirror the examples provided, rather than generating more generalizable responses.\qinyuan{Are you trying to compare one-shot cot with zero-shot cot in this sentence here?}\risha{I am comparing with directions vs without directions}
Interestingly, models such as \texttt{Llama-3.1}, \texttt{Med42}, and \texttt{Phi-3} exhibit moderate variation in performance on \hazmat{} data, suggesting that structured prompting can enhance accuracy \qinyuan{which ones are ``structured prompting''?}\risha{MF, Inchi, Iupac Name, SMILES}, but the extent of this benefit varies by model. These findings underscore the importance of tailoring prompt strategies to the specific architecture and reasoning capabilities of each model.

Insights from this pilot study informed our decision in Section~4.1 to adopt one-shot Chain-of-Thought (CoT) prompting for the structured translation task. This approach served as a balanced middle ground among prompt types, providing sufficient contextual guidance without excessive complexity. Alternatively, for the unstructured translation task in our main experiments, we employed zero-shot prompting. 
This choice was motivated by the substantial variation between \commonname and \synonyms, where prior examples offer little insight into representation patterns, unlike in the structured translation setting. \qinyuan{Do we have experiment results to support that?} \risha{This is just based on the definitions of the representation types. Common name and especially Un number given you little to no context about the chemical's breakdown }


\begin{figure}[ht]
    \centering
    \includegraphics[width=0.8\textwidth]{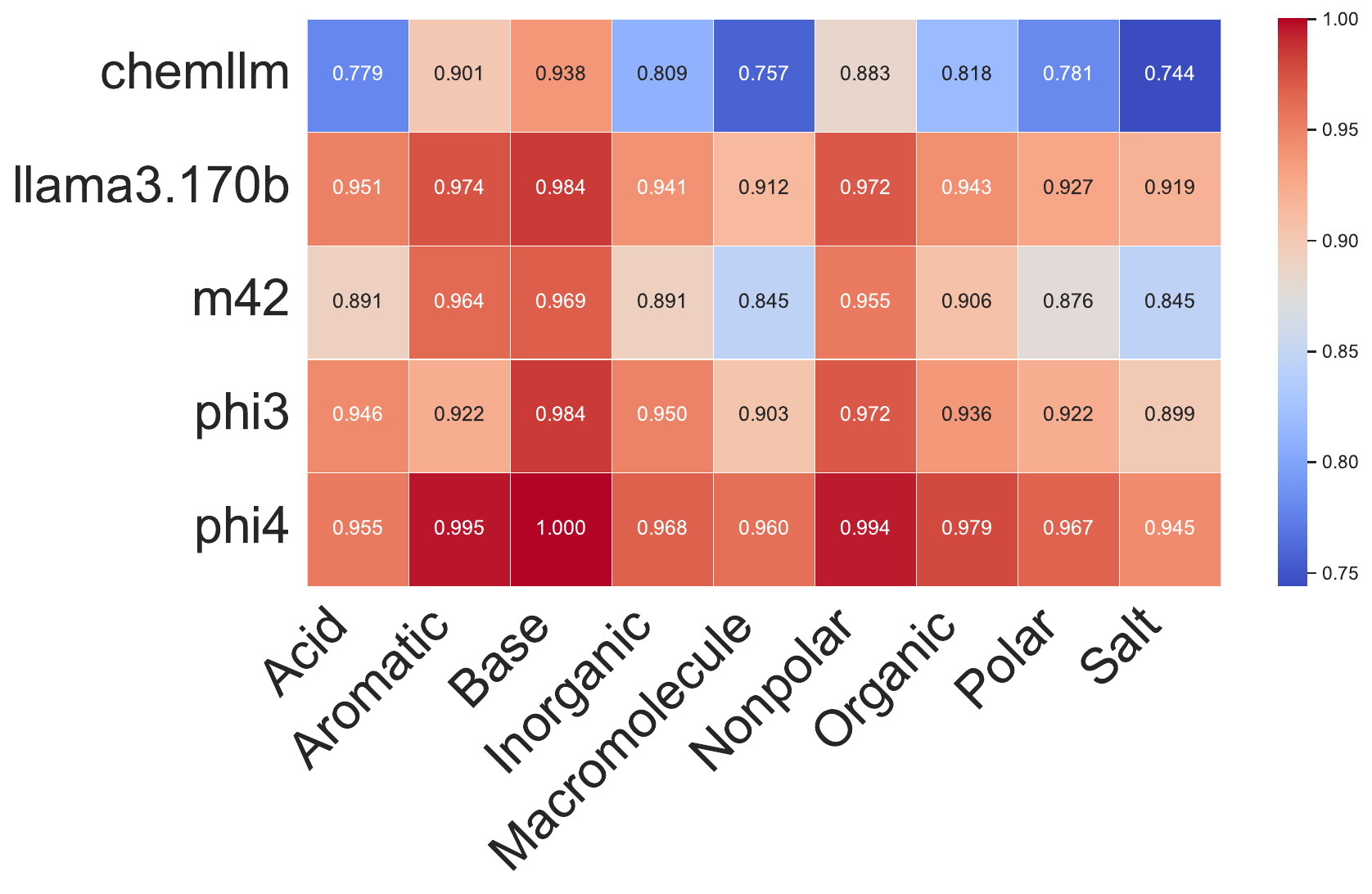}
    \caption{\textbf{Exact Match Scores for Model Performance on \texttt{hazmat} Data Across Chemical Categories.} 
    This heatmap illustrates how different models perform when translating hazardous material data, highlighting variations in accuracy across chemical types.}
    \label{fig:hazmat_heatmap_molecule}
\end{figure}


\subsubsection{Chemical Properties} 
The provided heatmaps in \autoref{fig:hazmat_heatmap_molecule} compare the performance of various models across nine distinct chemical categories: Acid, Aromatic, Base, Inorganic, Macromolecule, Nonpolar, Organic, Polar, and Salt. 

Phi-4 consistently outperforms the other models, achieving nearly perfect performance in the Base category and maintaining high accuracy across the remaining categories. ChemLLM lags behind the other models, particularly in the Inorganic, Nonpolar, and Salt categories, with scores generally below 0.81. While it performs well for simpler categories like Aromatic and Base, it struggles with more complex chemical types.

Certain categories display clear performance patterns across models. The \textbf{Base} and \textbf{Aromatic} categories achieve consistently high exact match (EM) scores, with five-model averages of 0.975 and 0.951, respectively, indicating that these chemical types are reliably easier for models to translate. In contrast, the \textbf{Inorganic} and \textbf{Nonpolar} categories present greater challenges, with lower average EM scores of 0.912 and 0.876, and notable drops in models like ChemLLM (0.809 and 0.757, respectively). Categories such as \textbf{Macromolecule} and \textbf{Polar} demonstrate moderate variability, with EM scores ranging from 0.757 to 0.960 for Macromolecule and 0.818 to 0.979 for Polar. These variations highlight that while some chemical representations are universally well-handled, others reveal weaknesses in certain models, particularly lower-performing ones like ChemLLM and Med42. Overall, this suggests that model architecture and training significantly influence performance on more complex or diverse chemical categories.\qinyuan{How do we arrive at this conclusion? How are ChemLLM and Med42 different from Phi-4 in their architecture?} \risha{domain-specific training?}


\begin{figure}[t]
    \centering
    \begin{subfigure}{0.48\linewidth}
        \centering
        \includegraphics[width=\linewidth]{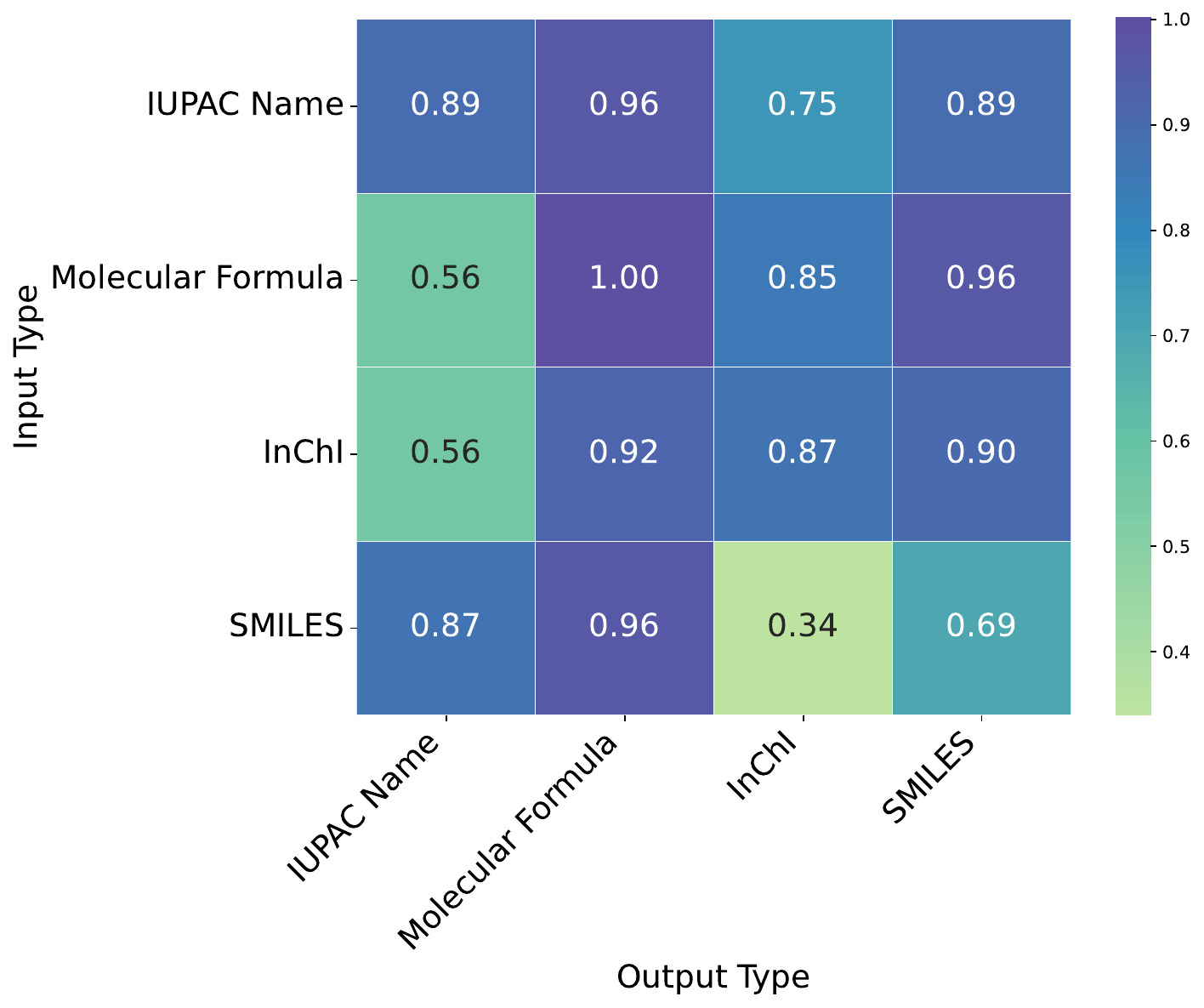}
        \caption{ChemLLM}
        \label{fig:chemllm}
    \end{subfigure}
    \hfill
    \begin{subfigure}{0.48\linewidth}
        \centering
        \includegraphics[width=\linewidth]{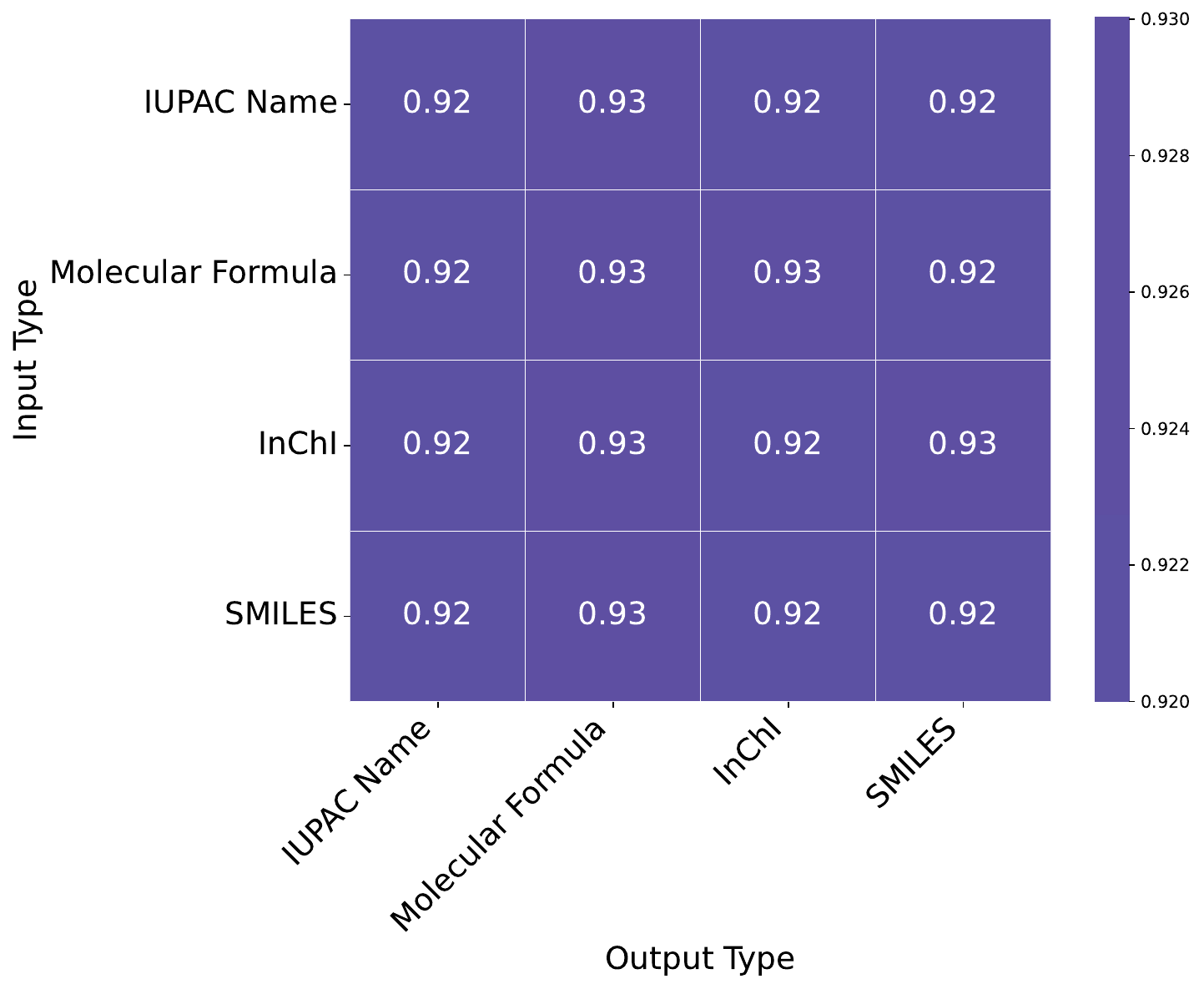}
        \caption{GPT-4o}
        \label{fig:gpt4o}
    \end{subfigure}
    
    \vspace{0.3cm}
    \begin{subfigure}{0.48\linewidth}
        \centering
        \includegraphics[width=\linewidth]{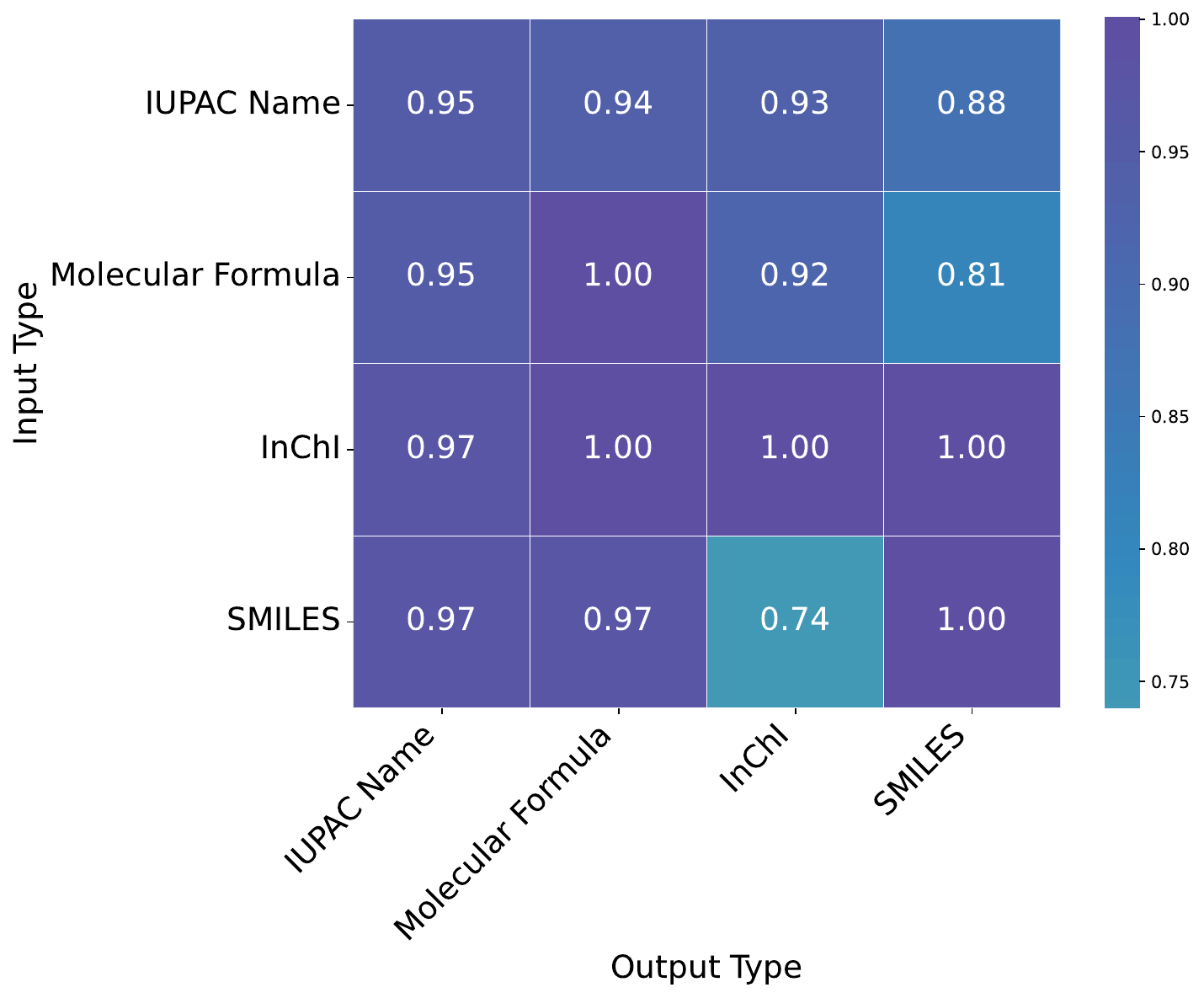}
        \caption{LLaMA 3.1}
        \label{fig:llama3}
    \end{subfigure}
    \hfill
    \begin{subfigure}{0.48\linewidth}
        \centering
        \includegraphics[width=\linewidth]{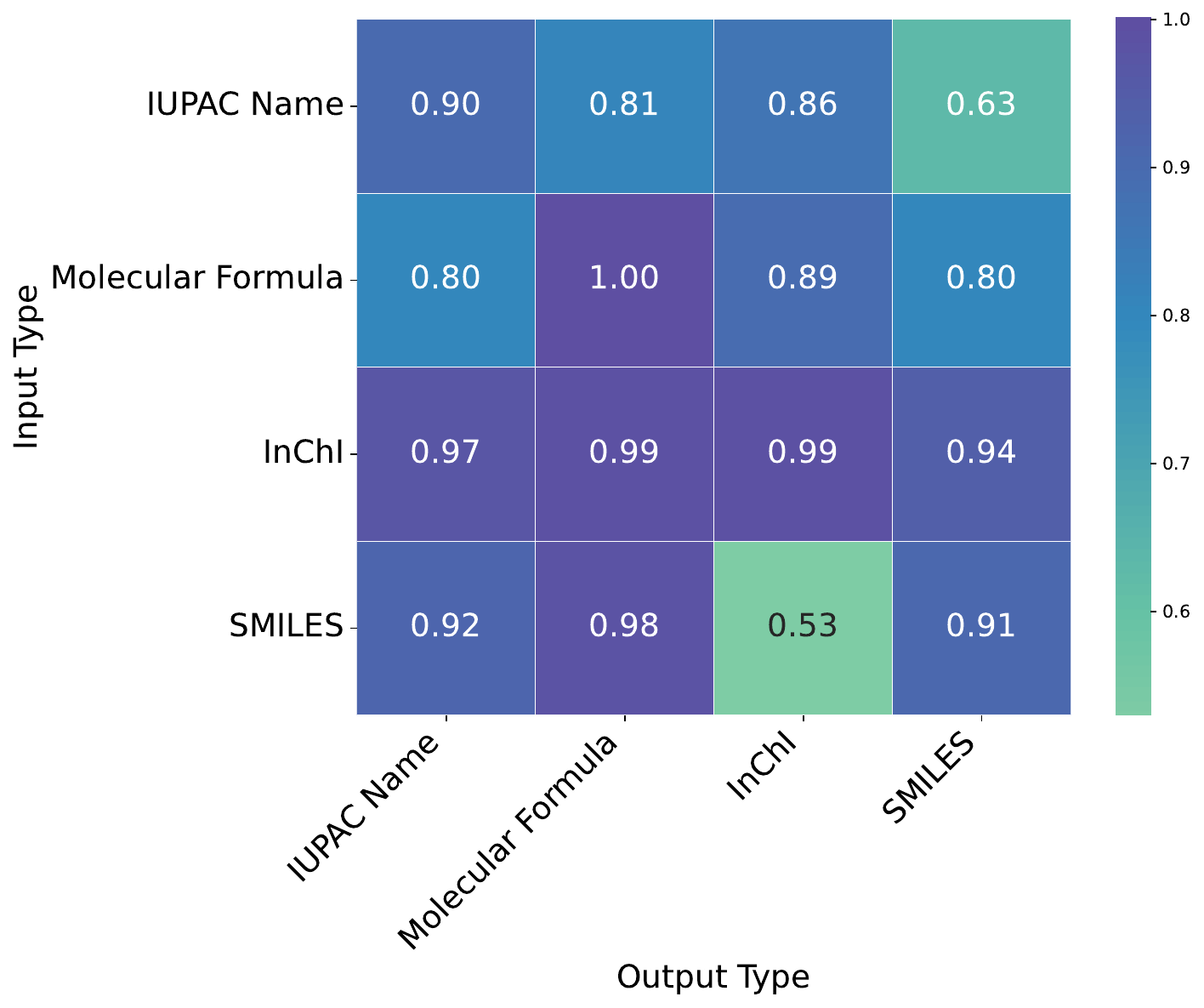}
        \caption{Med42}
        \label{fig:med42}
    \end{subfigure}
    
    \vspace{0.3cm}
    \begin{subfigure}{0.48\linewidth}
        \centering
        \includegraphics[width=\linewidth]{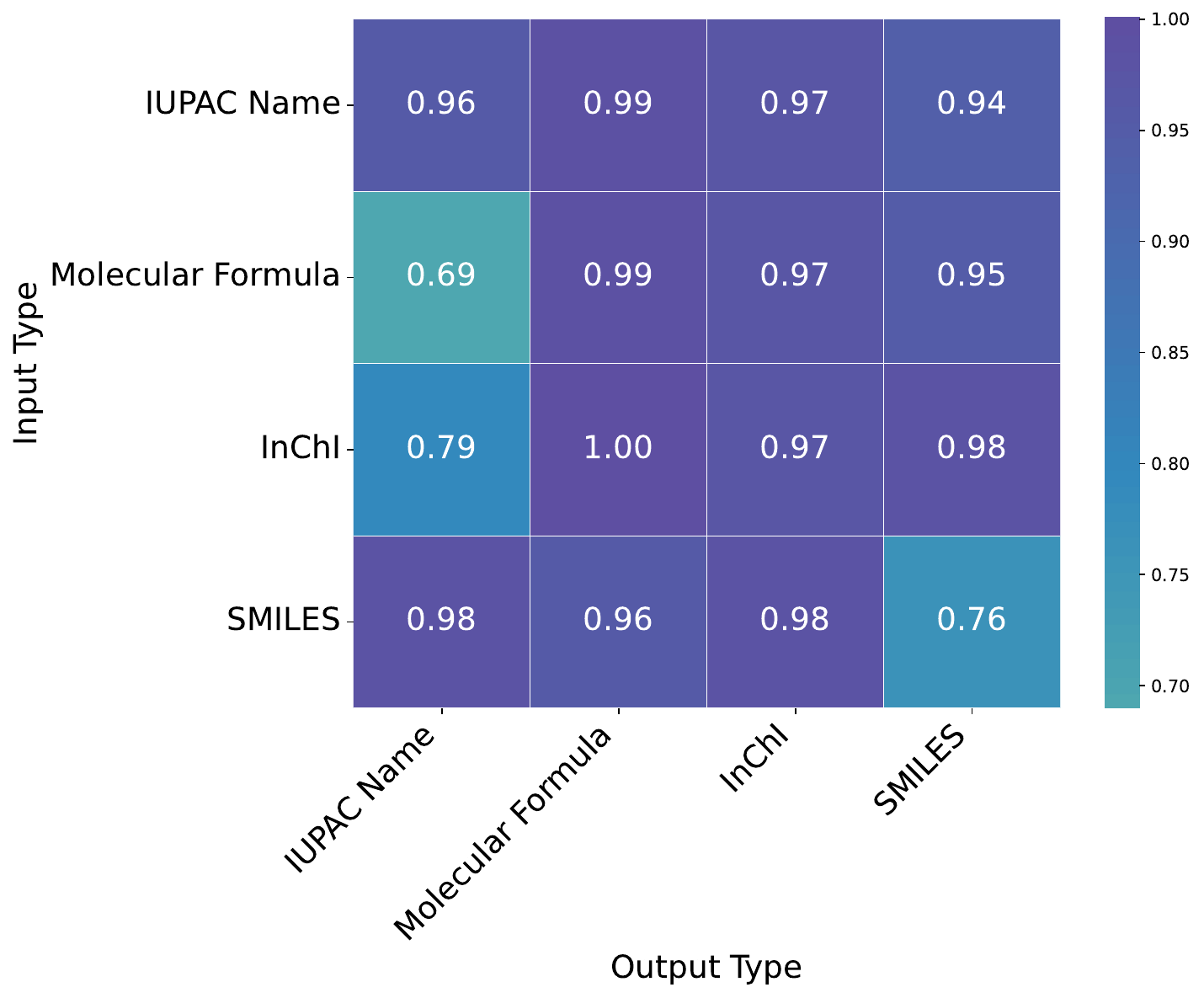}
        \caption{Phi-3}
        \label{fig:phi3}
    \end{subfigure}
    \hfill
    \begin{subfigure}{0.48\linewidth}
        \centering
        \includegraphics[width=\linewidth]{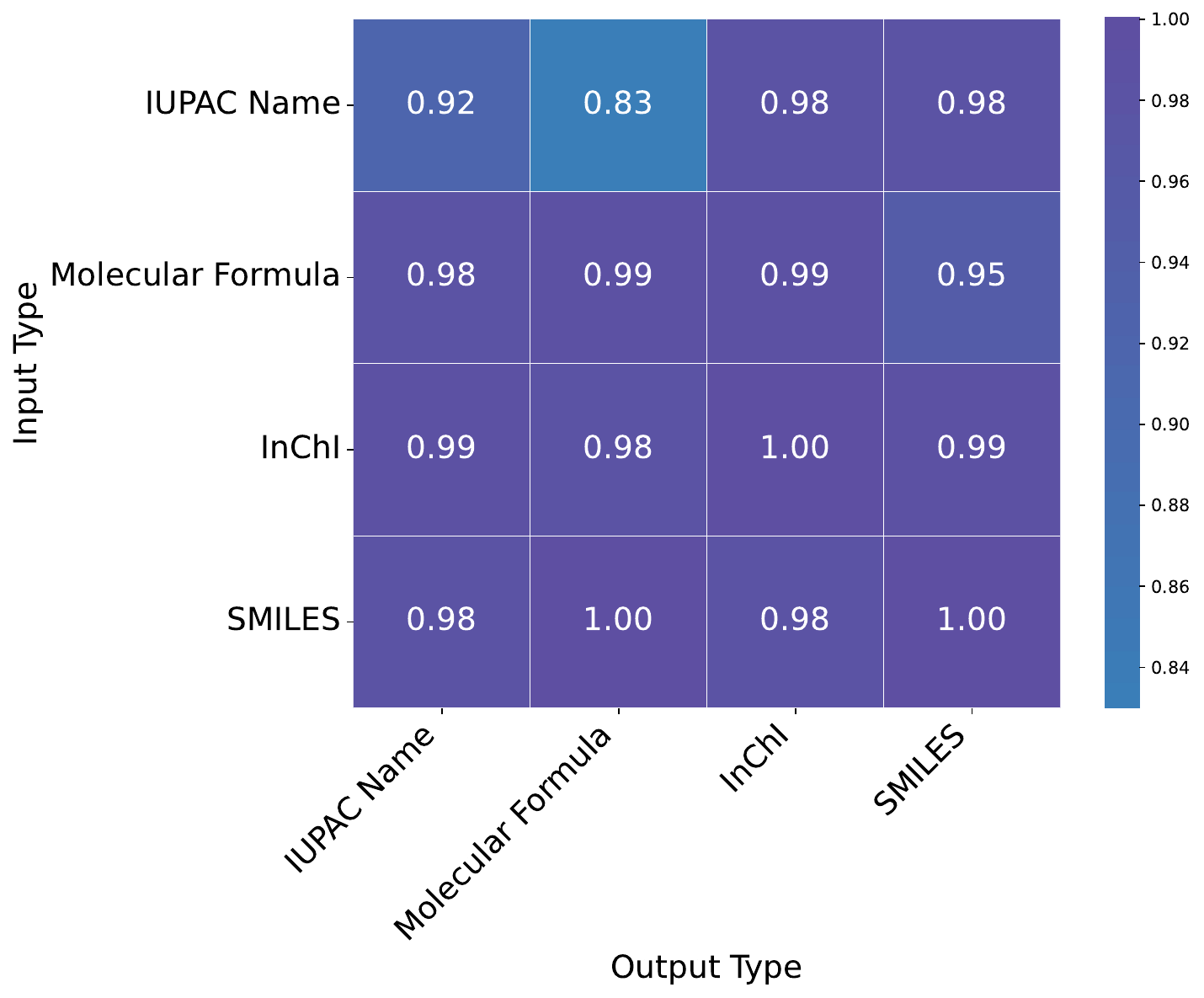}
        \caption{Phi-4}
        \label{fig:phi4}
    \end{subfigure}
    
    \caption{\textbf{Average Translation Task EM \texttt{hazmat} Scores Across Input/Output Representation Type.}
    This figure presents Exact Match (EM) scores for the Translation Task, grouped by model.}
    \label{fig:trans_var_main_result}
\end{figure}


\subsubsection{Input-Output Types}
\label{appendex:input-output-types}


\paragraph{Translation performance rarely varies with source chemical representation type for structured representation types.}
\autoref{fig:trans_var_main_result} provides a overall summary of all structured input-output combinations. For \textsc{ChemLLM}, we see that that an input-output combination of \smiles to \inchi is the most difficult translation. 

\paragraph{Translation performance significantly varies with source chemical representation type for unstructured representation types.}
Although \commonname and \unnumber are the only representations provided in the ERG, we observe that most models struggle to accurately identify a \unnumber and translate from a \commonname. Comparing the worst and best performing models, \textsc{Phi-3} and \textsc{GPT-4o}, respectively, we see that both models perform better when tasked with translating to a \commonname, which is encouraging for potential use by first responders. However, when translating away from a \commonname, models tend to perform better when translating to structured forms like \iupacname or \molecularformula rather than to a \unnumber. Despite this, performance when translating away from \commonname remains significantly lower, achieving only about half the accuracy observed when translating between structured forms, where models exhibit near-perfect performance.

Interestingly, the high translation scores on \synonyms suggest that models are more adept at recognizing alternative names for the same chemical, providing additional motivation for their use by first responders. Synonym recognition can be critical in high-stakes situations where rapid identification of hazardous materials is essential. By leveraging this strength, models can assist emergency personnel in identifying chemicals accurately, even when encountering less familiar terms, ultimately enhancing response efficiency and safety.

\begin{figure}[t]
    \centering
    \begin{minipage}{0.48\textwidth}
        \centering
        \includegraphics[width=\linewidth]{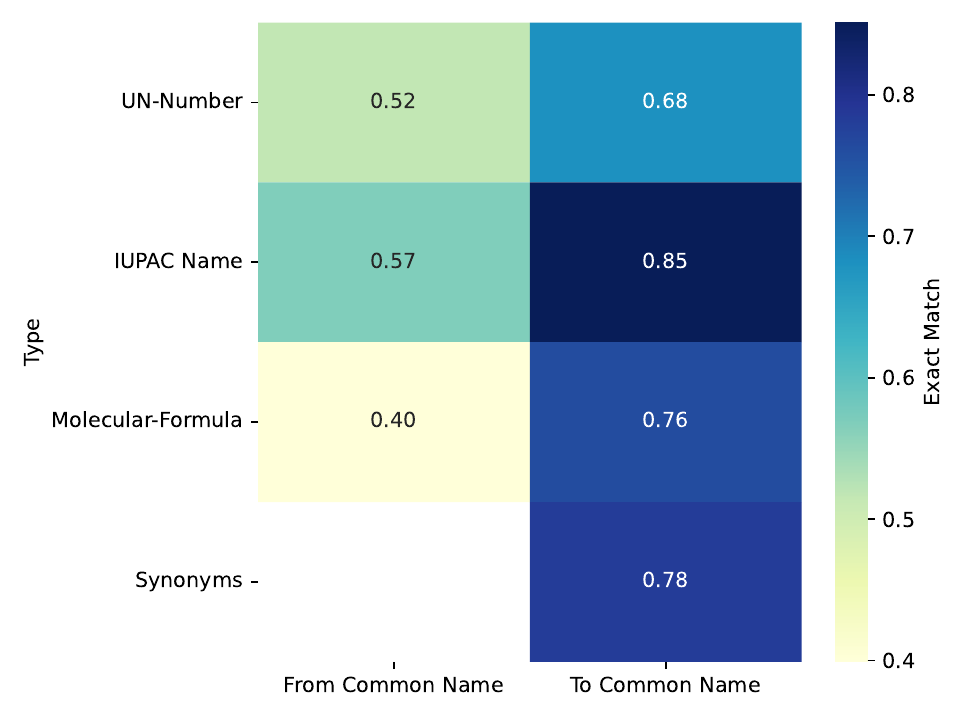}
        \caption{Exact Match Heatmap: Best Performing Model (GPT-4o).}
        \label{fig:heatmap_gpt4o}
    \end{minipage}
    \hfill
    \begin{minipage}{0.48\textwidth}
        \centering
        \includegraphics[width=\linewidth]{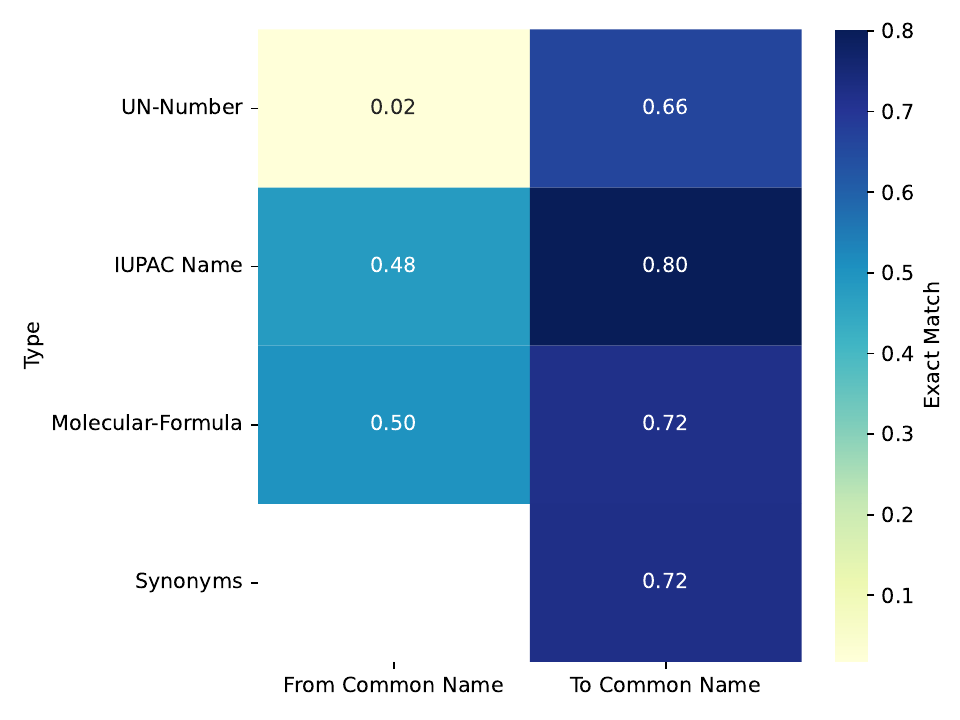}
        \caption{Exact Match Heatmap: Worst Performing Model (Phi-3).}
        \label{fig:heatmap_chemllm}
    \end{minipage}
\end{figure}



\subsubsection{\hazmat Chemicals v.s. Non-\hazmat chemicals.}
\paragraph{Translation of more common \hazmat chemicals is easier than non-\hazmat chemicals.}
In this section, we compare Chemical Familiarity by \hazmat vs Non-\hazmat data. Non-\hazmat Data was acquired from the PubChem database, and a subset of 100 chemicals were selected.



\begin{figure}[t]
    \centering
    \begin{minipage}{\textwidth}
        \centering
        \includegraphics[width=\textwidth]{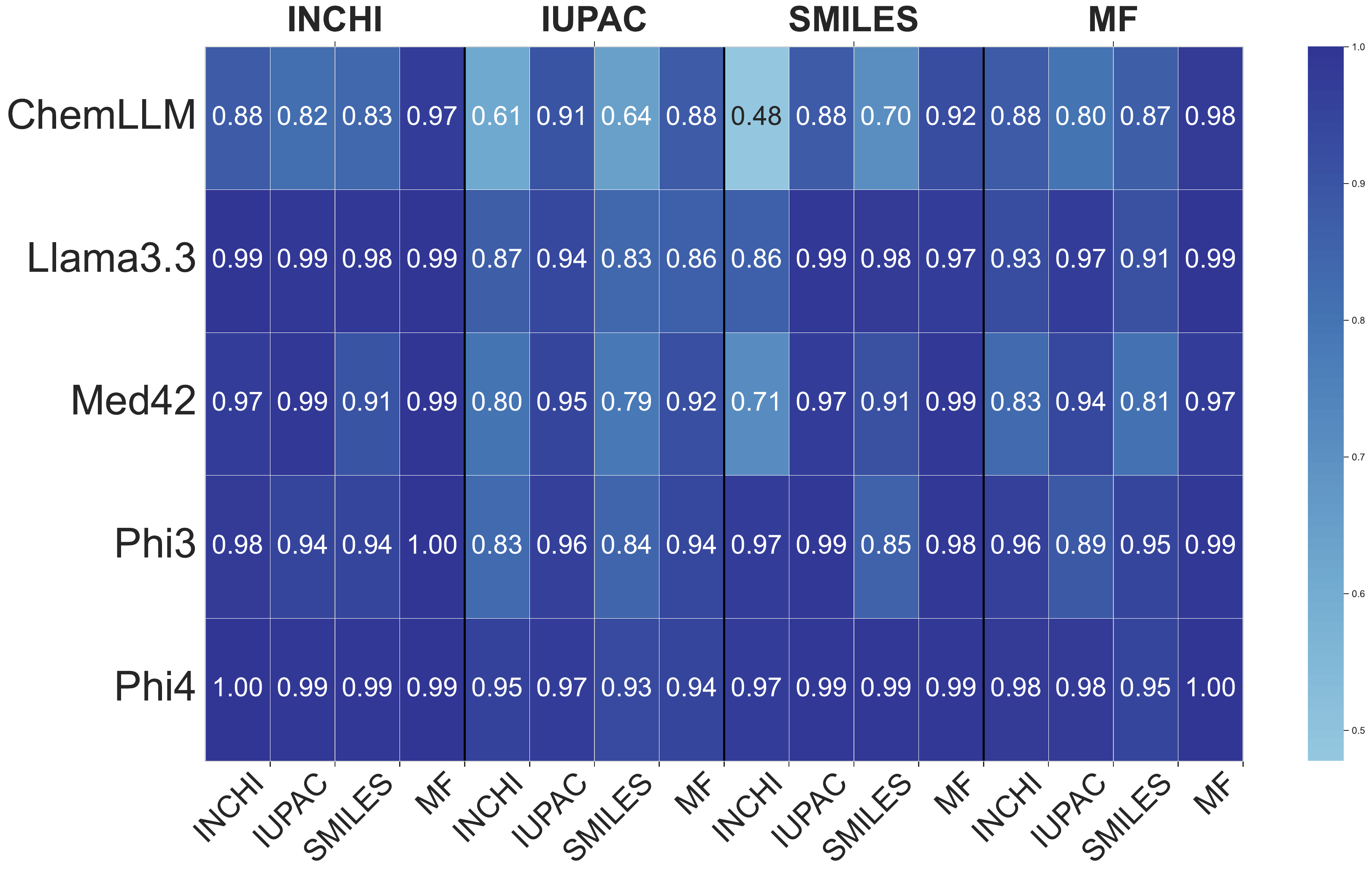}
        \caption{\textbf{EM Averages for \hazmat Data across Input and Output Representation Types.} 
        Each heatmap cell corresponds to a translation task from an \textbf{input type} (top row) to an \textbf{output type} (bottom column). 
        Note that these results are based on a different evaluation cohort (100 distinct chemicals) compared to Fig.~\ref{fig:trans_var_main_result}, leading to variation in EM scores.
        }

        \label{fig:supplementary_trans_hazmat}
    \end{minipage}
    
    \vspace{0.5cm}
    
    \begin{minipage}{\textwidth}
        \centering
        \includegraphics[width=\textwidth]{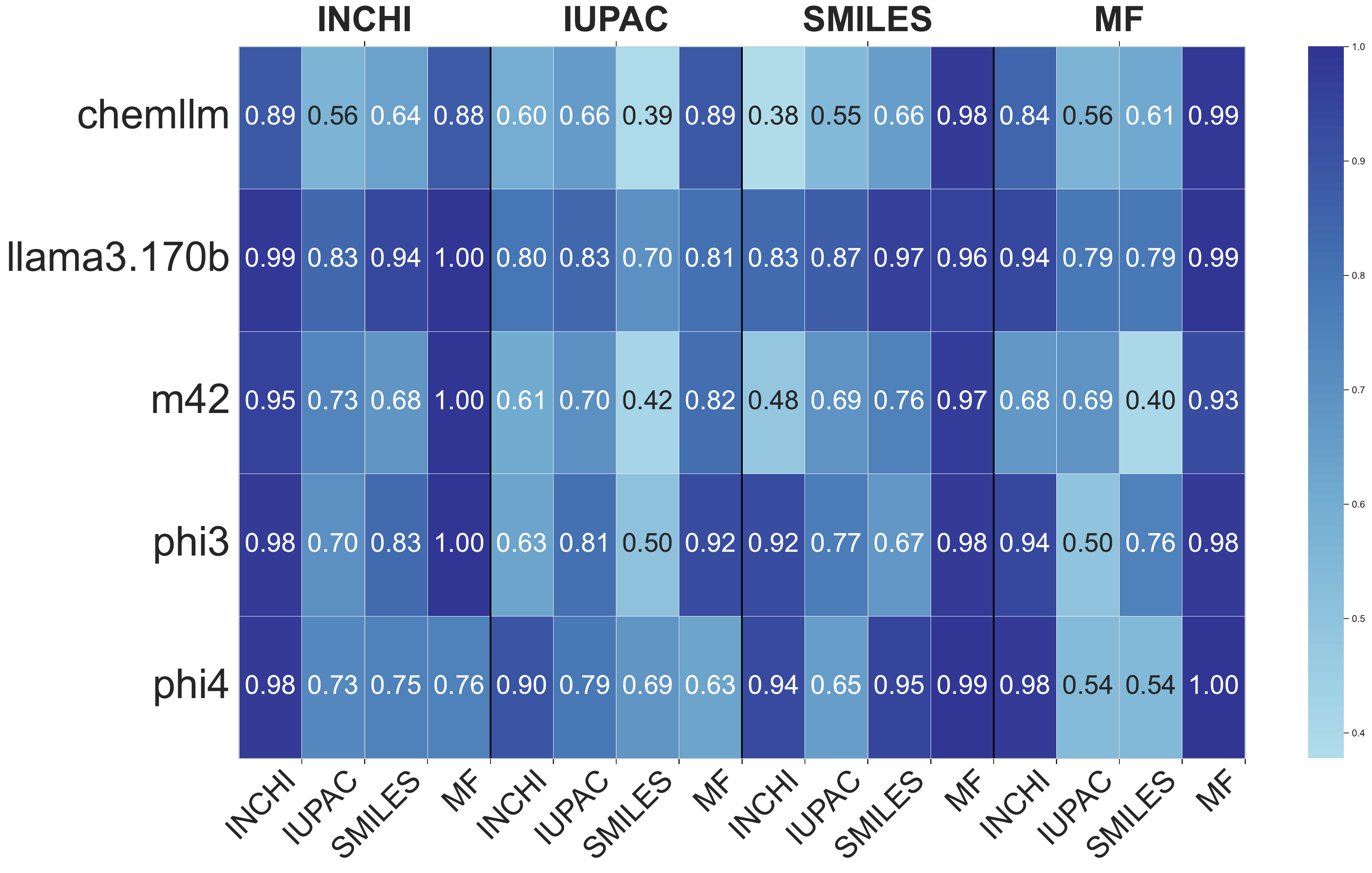}
        \caption{\textbf{EM Averages for Non-\hazmat Data across Input and Output Representation Types.} 
        Similar to Fig.~\ref{fig:supplementary_trans_hazmat}, this heatmap reflects translations between input (top row) and output (bottom column) types using a separate cohort of PubChem chemicals.}
        \label{fig:supplementary_trans_pub}
    \end{minipage}
\end{figure}

Overall, in \autoref{fig:supplementary_trans_hazmat}, the Translation Task on \hazmat data received a 90.9\%, whereas the Translation task on Non-\hazmat data scored a 77.7\%. 
The significant improvement in chemical comprehension and translation accuracy are encouraging towards the use of LLMs for \hazmat scenarios, because they suggest that general-models have more knowledge on \hazmat data than that of the non-\hazmat corpus.
The improved performance on \hazmat data suggests that even general-purpose models can effectively capture the complexities of hazardous materials—an essential capability for safety-critical applications like emergency response, chemical spill mitigation, and regulatory oversight.

\subsection{Analyzing Task II: Incident Response}


\subsubsection{Chemical Guides}
\begin{figure}[t]
    \centering
    \includegraphics[width=\columnwidth]{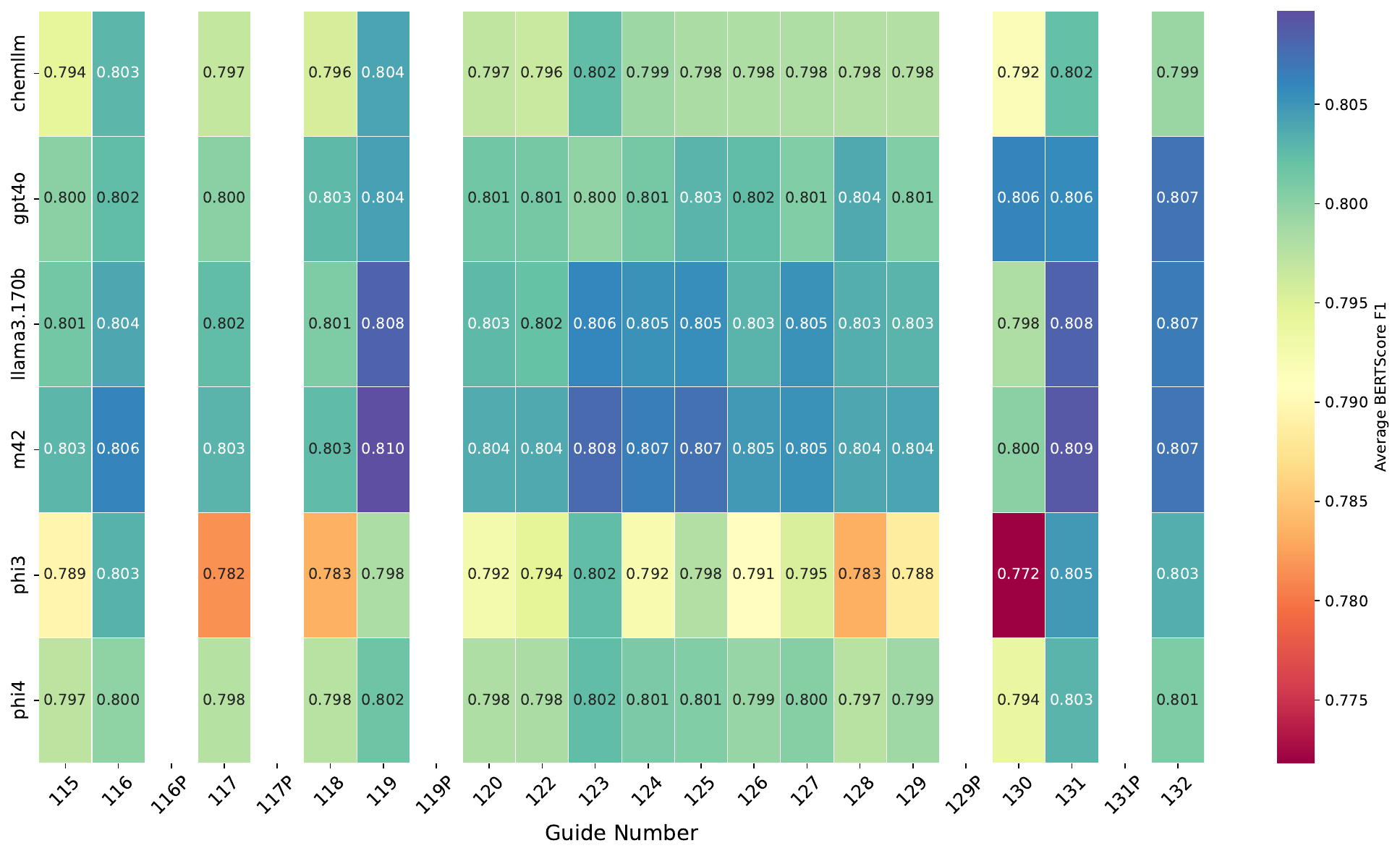}
    \caption{\textbf{Translation Task Model Performance:} Evaluation Across Molecule Types.\qinyuan{This is task 2 results right? The caption says Translation Task right now.} 
    \risha{This is for the Translation Task! we are simply looking at the results on the translation task, split by chemical guide (many chemicals share guides, which helps group them into similar categories and provide context into what a model is better at).} This figure illustrates how model performance varies when translating chemical representations for different molecule types. Each \hazmat molecule is associated with a specific guide number, which categorizes chemicals based on shared safety characteristics such as flammability, toxicity, and reactivity. Since multiple chemicals can correspond to the same guide number, these guides serve as a standardized way to group similar substances, providing a structured framework for understanding performance variations across molecular categories.}
    \label{fig:guide_number_main_result}
\end{figure}

Observing ERG Guide numbers allow us to pinpoint performance by chemical properties.
Each \hazmat chemical is linked to one of hundreds of guides in the ERG, explaining what to do in case of a chemical emergency. There can be more than one chemical linked to the same ERG guide, as many substances share similar hazards and response protocols. These guides provide critical information on fire and explosion risks, health hazards, protective measures, spill containment, and first aid procedures. Emergency responders rely on them to determine safe evacuation distances, recommended PPE, and appropriate mitigation strategies. Given the vast number of chemicals and their potential dangers, efficiently associating a chemical with the correct ERG guide is essential for rapid and effective decision-making in emergency situations. In \autoref{fig:guide_number_main_result}, we can see that there is a fairly even distribution across guides (between 77.5\% - 80.5\%) in terms of most answered correctly and incorrectly, suggesting that models may not be partial to solely flammable or toxic chemicals. 

\begin{wrapfigure}{r}{0.49\textwidth}
    \centering
    \includegraphics[width=\linewidth]{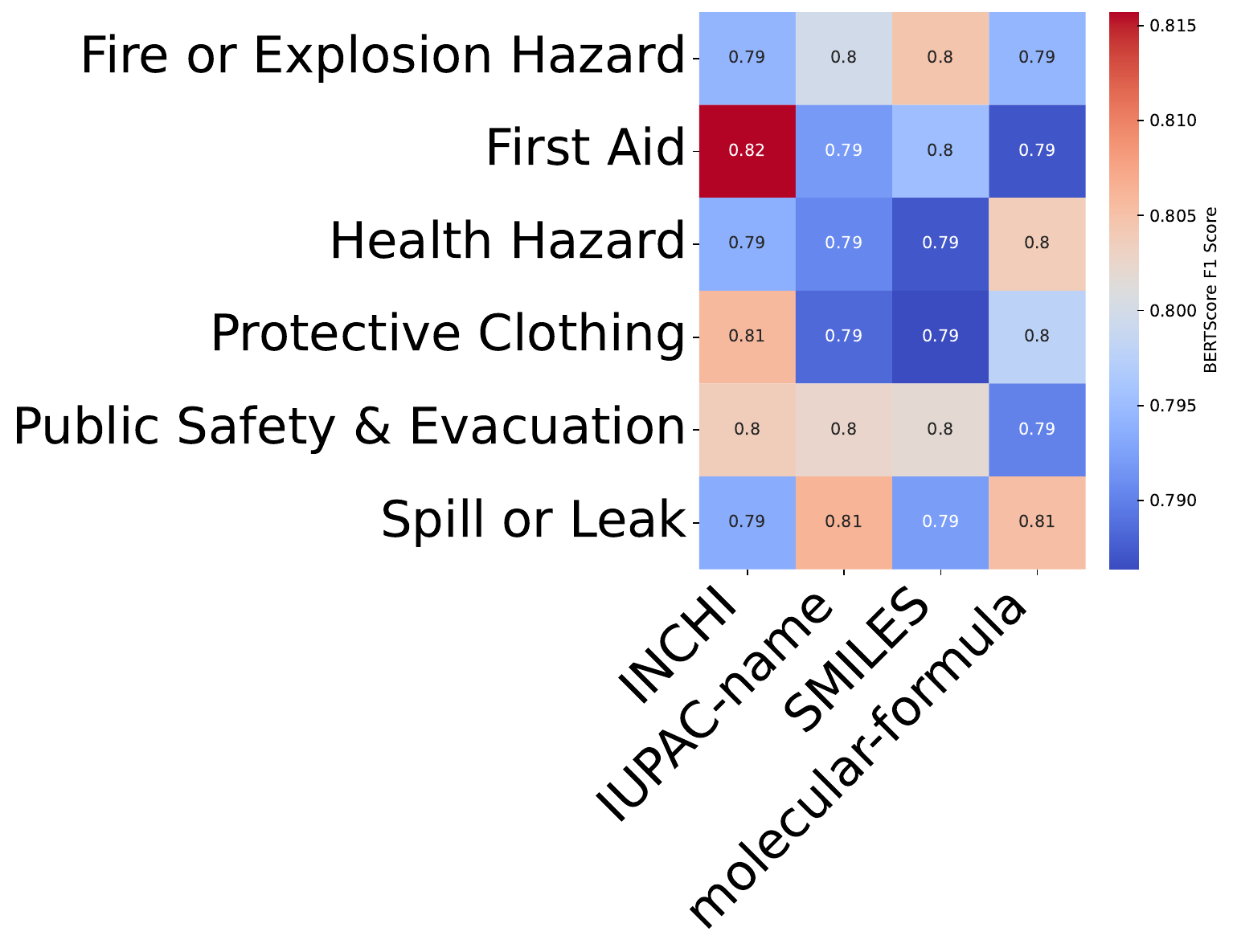}
    \caption{\textbf{Incident Response Task Model Performance:} Evaluation Using BERTScore-F1, average scores across all models.
    \qinyuan{Is this figure discussed somewhere?}
    }
    \label{fig:bertscore_results}
\end{wrapfigure}

\subsubsection{Models Often Provide Additional Information Beyond ERG Guidance}
\label{appendix:t2_model_outputs}
As noted in the main text, models frequently generate responses that go beyond the explicit guidance in the Emergency Response Guidebook (ERG), often including supplementary safety recommendations. Upon manually analyzing a subset of 20 model outputs from \textsc{GPT-4o}, we observed that the models tend to generate extended responses that incorporate multiple safety protocols, regardless of the specific prompt category.

For instance, The following example illustrates how when prompted with a query about \textit{Fire or Explosion Hazard} for Sodium superoxide, GPT-4o not only provided fire suppression guidance but also included additional considerations such as evacuation distances, personal protective equipment (PPE), and hazard assessment protocols. Despite the prompt explicitly asking for information relevant to fire or explosion hazards, the model integrated relevant details from other ERG guide subcategories, demonstrating an implicit understanding of the intersectional nature of hazardous material incidents.
\begin{lstlisting}[style=mystyle]
In dealing with a fire involving Sodium superoxide (NaO2), it is essential to consider its chemical properties and associated hazards. Sodium superoxide is a powerful oxidizer and can react vigorously with combustible materials, leading to potential fire and explosion risks. Here are my recommendations:

### Fire Control
1. **Use of Appropriate Extinguishing Agents:**
   - **Do Not Use Water:** Sodium superoxide reacts with water to produce sodium hydroxide and oxygen gas, which can exacerbate the fire. Moreover, the heat generated during this reaction can cause further thermal decomposition, releasing more oxygen.
   - **Avoid Carbon Dioxide and Foam:** These are ineffective because sodium superoxide can decompose them, potentially leading to further reactions.
   - **Use Dry Powder Extinguishers:** Class D fire extinguishers (specifically designed for metal fires) containing dry powder agents such as sodium chloride or graphite are recommended. These agents can smother the fire without providing additional fuel (oxygen or moisture).

2. **Isolation of Fuel Sources:**
   - Remove or isolate any nearby combustible materials to prevent further reaction and fire spread.

### Explosion Risks
1. **Reactivity with Combustibles:** Sodium superoxide can react explosively with organic materials (e.g., paper, wood) and certain metals. Ensure that such materials are not present in the vicinity or are removed promptly.
2. **Controlled Ventilation:**
   - Ensure adequate ventilation to prevent the accumulation of oxygen-rich atmospheres that can increase the risk of explosion.
3. **Temperature Control:**
   - Cool surrounding areas with inert gases (if available) or non-reactive cooling methods to prevent the decomposition of sodium superoxide, which releases oxygen.

### Evacuation Procedures
1. **Initial Evacuation Radius:**
   - Establish an initial evacuation perimeter of at least 150 meters (500 feet) to account for potential explosions and toxic smoke.
2. **Hazard Assessment:**
   - Conduct a continuous hazard assessment to determine the need for expanding the evacuation zone. Factors include the fire's intensity, wind direction, and presence of hazardous materials.
3. **Personnel Protection:**
   - Ensure that all responders and personnel in the vicinity are equipped with appropriate personal protective equipment (PPE), including self-contained breathing apparatus (SCBA) to protect against inhalation of toxic fumes.
4. **Communication:**
   - Maintain clear and constant communication with emergency personnel and responders about the status of the fire.
\end{lstlisting}

\paragraph{Evaluation Summary}
The model output provides a detailed and accurate description of the fire and explosion hazards associated with Sodium superoxide, including its reactivity with combustibles and the need for appropriate extinguishing agents. It also covers the potential for explosive reactions and the importance of controlling ventilation and temperature. However, it does not explicitly mention the possibility of containers exploding when heated or the risk of runoff creating a fire or explosion hazard, which are key points in the ERG guidance. This highlights how models tend to extrapolate relevant safety protocols even when not explicitly requested. The LLM Judge, scored this a Partial (0.5).

\subsection{Human Evaluations}
\label{appendix:t2-results}
\paragraph{Human evaluation by a first responder shows alignment with the LLM-Judge.}  
We emphasize that Tasks 1 and 3 have verifiable answers that can be evaluated deterministically: Task~1 uses Exact/Partial Match on chemical representations, while Task~3 relies on Multiple Choice Answer Keys. Only Task~2 requires long-form responses, where the gold-standard outputs were not LLM-generated but instead obtained directly from the federally standardized \emph{Emergency Response Guidebook} (ERG), the authoritative reference for emergency responders.

For Task~2, we adopt a combination of Exact Match, BERTScore, and LLM-based judgment as a starting point. We acknowledge, however, that these automated metrics cannot fully substitute for human validation. In particular, using GPT-4 as a judge when GPT-4o is one of the evaluated models raises potential bias concerns. To mitigate this, one of the authors—a certified EMT with HAZMAT training and five years of field experience—conducted multiple rounds of testing to refine and validate outputs. While large-scale human validation was beyond the scope of this initial study, we consider it a critical next step as we expand to more complex emergency scenarios.

The LLM Judge score measures alignment between a model’s response and the ERG reference. For example, a score of 52.7\% indicates that while models can capture the broad contours of appropriate guidance (e.g., recognizing that PPE is required), they often lack specificity or completeness (e.g., omitting gloves or eye protection). Such partial correctness can present real-world risks, underscoring the importance of human oversight for emergency response applications.

To further probe reliability, we conducted a small-scale human validation study of Task~2 responses. We sampled 96 responses across six prompt types, focusing on the top two performing models and the lowest performing model. Both the LLM-Judge and a HAZMAT-certified EMT independently reviewed each prompt, model output, and corresponding ERG section. Additional relevant but accurate information was accepted, while omissions of significant ERG content or introduction of inaccuracies were penalized.  

\begin{table}[h]
\centering
\begin{tabular}{lccc}
\toprule
Model & Agreement & LLM-Annotator Avg (\%) & Human Avg (\%) \\
\midrule
Llama3 & 88.5 & 69.8 & 62.0 \\
Phi4   & 94.8 & 86.5 & 84.4 \\
GPT4o  & 76.0 & 69.3 & 68.8 \\
\bottomrule
\end{tabular}
\caption{Comparison of Task~2 evaluation between LLM-Judge and human validator (HAZMAT-certified EMT with 5 years of experience).}
\end{table}

\section{Analyzing Task III: \hazmat Examination}

\paragraph{Contamination check.} To ensure that model performance reflects understanding and generalization instead of memorization, we conducted exact-match searches using the Infini-gram engine \citep{Liu2024InfiniGram} over several open pretraining corpora—\texttt{Dolma-v1.7} \citep{soldaini-etal-2024-dolma}, \texttt{RedPajama} \citep{weber2024redpajama}, \texttt{Pile-train} \citep{gao2020pile}, and \texttt{C4-train} \citep{2020t5}. No overlaps were found with our multiple-choice QA dataset.

\end{document}